\documentclass{article}

    \usepackage[preprint]{neurips_data_2024}
\usepackage[utf8]{inputenc} % allow utf-8 input
\usepackage[T1]{fontenc}    % use 8-bit T1 fonts
\usepackage{hyperref}       % hyperlinks
\usepackage{xcolor}
\usepackage{url}            % simple URL typesetting
\usepackage{booktabs}       % professional-quality tables
\usepackage{amsfonts}       % blackboard math symbols
\usepackage{nicefrac}       % compact symbols for 1/2, etc.
\usepackage{microtype}      % microtypography
\usepackage{xcolor}         % colors
\usepackage{paralist}
\usepackage{multirow}
\usepackage{amsmath}
\usepackage{graphicx}
\usepackage{subcaption}
\usepackage{tikz}
\usepackage{wrapfig}
\usepackage{colortbl}

\title{Learning on Graphs with Large Language Models (LLMs): A Deep Dive into Model Robustness}

\author{%
Kai Guo$^1$\thanks{Equal Contribution}\,,  Zewen Liu$^{2\ast}$\,, Zhikai Chen$^3$\,, Hongzhi Wen$^3$ \\ 
\textbf{Wei Jin$^2$\,, Jiliang Tang$^3$\,, Yi Chang$^1$} \\
 $^1$Jilin University, $^2$Emory University,$^3$Michigan State University\\
  \texttt{guokai20@mails.jlu.edu.cn,}
 \texttt{\{zewen.liu,wei.jin\}@emory.edu,}\\
  \texttt{\{chenzh85,wenhongz,tangjili\}@msu.edu,}
\texttt{yichang@jlu.edu.cn}
}

\begin{document}

\maketitle

\begin{abstract}
Large Language Models (LLMs) have demonstrated remarkable performance across various natural language processing tasks. Recently, several LLMs-based pipelines have been developed to enhance learning on graphs with text attributes, showcasing promising performance. However, graphs are well-known to be susceptible to adversarial attacks and it remains unclear whether LLMs exhibit robustness in learning on graphs. To address this gap, our work aims to explore the potential of LLMs in the context of adversarial attacks on graphs. Specifically, we investigate the robustness against graph structural and textual perturbations in terms of two dimensions: LLMs-as-Enhancers and LLMs-as-Predictors. Through extensive experiments, we find that, compared to shallow models, both LLMs-as-Enhancers and LLMs-as-Predictors offer superior robustness against structural and textual attacks.
% while LLMs-as-Predictors demonstrate superior robustness against both structural attacks and textual attacks.  
Based on these findings, we carried out additional analyses to investigate the underlying causes. Furthermore, we have made our benchmark library openly available to facilitate quick and fair evaluations, and to encourage ongoing innovative research in this field.
\end{abstract}
\section{Introduction}
\label{sec:intro}
In recent years, significant progress has been made in the development of Large Language Models (LLMs) like Sentence-BERT~\cite{reimers2019sentence}, GPT~\cite{radford2018improving}, LLaMA~\cite{touvron2023llama}, etc. These variants showcase exceptional performance across a range of natural language processing tasks, such as sentiment analysis~\cite{sun2023sentiment, ronningstad2024gpt}, machine translation~\cite{feng2024improving, zhang2023prompting}, and text classification~\cite{sun2023text,zhang2024generation}.  While LLMs are widely employed for handling plain text, there is an increasing trend of applications where text data is linked with structured information represented as text-attributed graphs (TAGs)~\cite{chen2024exploring,he2023harnessing}. Recently, solely utilizing LLMs for graph data has proven effective in various downstream graph-related tasks, and integrating LLMs with Graph Neural Networks (GNNs)~\cite{kipf,velivckovic2017graph} can further enhance graph learning capabilities~\cite{chen2024exploring}.

Although graph machine learning methods with LLMs (Graph-LLMs) have reported promising performance~\cite{qin2023disentangled,chen2023label,wei2024llmrec,guo2023gpt4graph,zhao2024gimlet,wang2024can,cao2023instructmol,liu2023git,qian2023can,chien2021node,duan2023simteg}, their robustness to adversarial attacks remains unknown.
Robustness has always been a crucial aspect of model performance, especially in high-risk tasks like medical diagnosis~\cite{ahmedt2021graph}, autonomous driving~\cite{xiao2023review}, epidemic modeling~\cite{liu2024review}, where failures can have severe consequences. It is well-known that graph learning models such as GNNs are vulnerable to adversarial attacks, where adversaries manipulate the graph structure to produce inaccurate predictions~\cite{zugner2018adversarial,jin2021adversarial}. Moreover, text attributes in TAGs are also at risk of manipulation by attackers, which further raises concerns about the reliability of graph learning algorithms in safety-critical applications. In the era of LLMs embracing graphs, a critical question arises: \textit{Are Graph-LLMs robust against graph adversarial attacks?}

\begin{figure}[t]
  \centering
  \label{fig:framework}
  \includegraphics[width=1\linewidth]{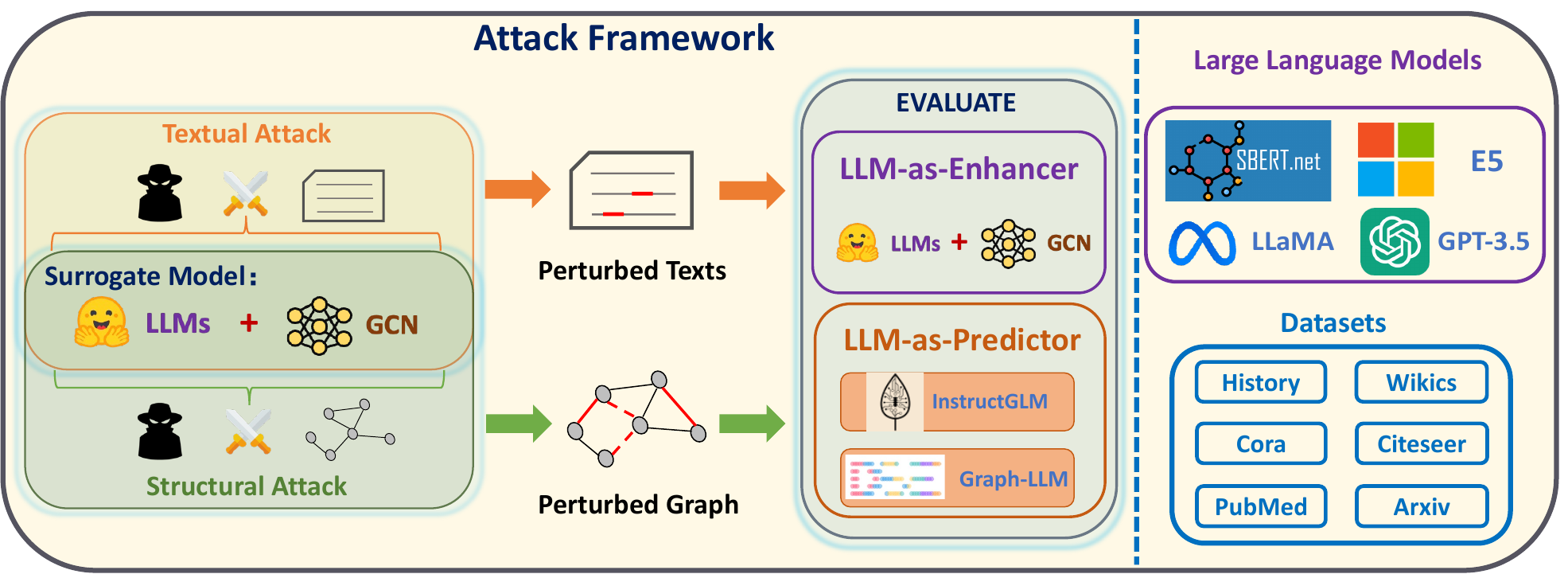} 
  \caption{An overview of our benchmark. The evaluation is divided into two perspectives: LLMs-as-Enhancers and LLMs-as-Predictors, both of which consider structural and textual attacks.}
  \vskip -1.9em
\end{figure}

To address this question, we identify several research gaps in existing evaluations that hinder our understanding of current methods. These include: (1) \textbf{Limited Structural Attacks for Graph-LLMs}: Existing structural attacks are tailored for GNNs and have not been tested on Graph-LLMs, leaving it uncertain whether Graph-LLMs are sensitive to subtle structural changes within graphs. Understanding this sensitivity is crucial for developing robust Graph-LLMs.
(2) \textbf{Limited Textual Attacks for TAGs}: Current feature attacks~\cite{zugner2020adversarial,ma2020towards} particularly manipulate node attributes in the embedding of continuous space, rather than in textual space. This calls for creating an evaluation framework for assessing the robustness of Graph-LLMs to textual attacks. This area has been minimally explored, and it is essential to understand how text manipulations within graphs can impact model performance. 
(3) \textbf{Diverse Architectures of Graph-LLMs}: Various strategies exist for utilizing LLMs for graph data, with approaches such as LLMs-as-Enhancers~\cite{chen2024exploring, he2023harnessing, zhao2022learning}and LLMs-as-Predictors~\cite{chen2024exploring, ye2023natural, chai2023graphllm} being among the most popular. This diversity necessitates the development of specialized attack pipelines to address the unique characteristics of different Graph-LLM architectures.

In response to these challenges, we aim to conduct a fair and reproducible evaluation of both structural and textual attacks under the representative pipelines of Graph-LLMs for node classification. Our contributions can be summarized  as follows:
\begin{compactenum}[\textbullet]
\item 
\textbf{New evaluation perspective.} Different from existing works focusing on the predictive power of Graph-LLMs, 
we stress test the robustness of Graph-LLMs against graph adversarial attacks. To the best of our knowledge, this problem has not been studied before. Our robustness assessment of each methodology includes both structural attack and textual attack aspects, while previous investigations on textual attacks on graphs are rather limited.

\item 
\textbf{Reproducible and comprehensive comparison.} 
We conduct a comprehensive comparison of various Graph-LLM pipelines across multiple datasets. Specifically, we systematically evaluate two distinct groups of Graph-LLMs: LLMs-as-Enhancers and LLMs-as-Predictors. To guarantee reproducibility and fairness in our comparison, we fine-tune all models using the same set of hyper-parameters. We employ two types of evaluation metrics: performance degradation percentage and attack success rate.

\item \textbf{Insights into the robustness of Graph-LLMs:} This study reveals several interesting observations about the robustness of Graph-LLMs: 
\begin{compactenum}[(a)]

\item 
LLMs-as-Enhancers exhibit greater robustness against adversarial \underline{structural attacks} compared to shallow embeddings like Bag of Words (BOW) and TF-IDF, particularly at high attack rates. Additionally, the better the distinguishability of the encoded features, the better the robustness.

\item 
LLMs-as-Predictors show better robustness in resisting \underline{structural attacks} than MLP and the fine-tuned predictor holds even better robustness than vanilla GNN. 

\item 
{LLMs-as-Enhancers} demonstrate excellent robustness against \underline{textual attacks}, with GCN being significantly more robust than MLP as the victim model. 

\item 
{Text entropy, text length, and node centrality show certain negative correlations with the attack success rate of the textual attack for Graph-LLMs.}

\end{compactenum}

\item 
\textbf{Open-source benchmark library:} To support and advance future research, we have developed an easy-to-use open-source benchmark library, now publicly available on GitHub: \url{https://github.com/KaiGuo20/GraphLLM_Robustness}. This library allows researchers to quickly evaluate their own methods or datasets with minimal effort. Additionally, we have outlined potential future directions based on our benchmark findings to inspire further investigations.

\end{compactenum}

{\section{Formulations and Background}}
\label{sec:background}
We begin by providing preliminaries on graph neural networks, and then formalize the graph adversarial attacks. Finally, we briefly introduce the developments in large language models on graphs. More details are shown in Appendix~\ref{sec:sup-background}.

\textbf{Notations.} We define a graph as $G = (V,E)$, where $V$ denotes the node set and $E$ represents the edge set. We employ $\mathbf{X} \in \mathbb{R}^{N \times d}$ to denote the node feature matrix, where $N$ is the number of nodes and 
$d$ is the dimension of the node features.  Furthermore, we use the matrix $\mathbf{A} \in \mathbb{R}^{N \times N}$ to signify the adjacency matrix of $G$. 
% In this matrix, each $\mathbf{A}_{ij}$ = 1 denotes a connection between nodes $v_i$ and $v_j$ in $G$. 
Finally, the graph data can be denoted as $G = (\mathbf{A},\mathbf{X})$. 

\textbf{Graph Neural Networks.} GCN~\cite{kipf} is one of the most representative models of GNNs, utilizing aggregation and transformation operations to model graph data. Unlike GCN, which treats all neighbors equally, GAT~\cite{velivckovic2017graph} assigns different weights to different nodes within a neighborhood during aggregation.

\textbf{Graph Adversarial Attacks.}
While graph adversarial attacks can perturb node features or graph structures, most existing attacks focus on modifying the graph structure due to its complexity and effectiveness. These modifications often involve adding, deleting, or rewiring edges~\cite{jin2020graph,madry2017towards,geisler2021robustness,xu2019topology,chang2020restricted,ma2019attacking,entezari2020all,chen2021time,zhang2021projective}, exemplified by the PGD~\cite{madry2017towards} and PRBCD\cite{geisler2021robustness} attacks. Besides evaluating structural attacks, we also adopt a more direct approach by using textual attacks instead of feature attacks targeting continuous features for evaluation, which remains a relatively unexplored direction.

\textbf{Textual Attack.}
Textual attacks can be performed on different levels like character level or sentence level according to the target to be perturbed. In this work, we focus on word-level attacks, applying substitutions to fool the classifier with minimal text perturbation. For example, \textbf{SemAttack} \cite{wang2022semattack} generates adversarial text by employing various semantic perturbation functions.

\textbf{Large Language Models (LLMs) on Graphs.}
Recent advances in Large Language Models (LLMs) like BERT~\cite{devlin2018bert}, Sentence-BERT (SBert)~\cite{reimers2019sentence} E5~\cite{wang2022text}, GPT~\cite{radford2018improving}, LLaMA~\cite{touvron2023llama} and their variants have significantly impacted graph-related tasks. Two main paradigms are LLMs-as-Enhancers, improving node features (e.g., TAPE~\cite{he2023harnessing}, KEA~\cite{chen2024exploring}, GLEM~\cite{zhao2022learning}), and LLMs-as-Predictors, leveraging LLMs for graph predictions (e.g., InstructGLM~\cite{ye2023natural}, GraphLLM~\cite{chai2023graphllm}, GraphGPT~\cite{tang2023graphgpt}).

\section{Benchmark Design}
\vspace{-0.5em}

\label{sec:design}
To deepen our understanding of the potential of Graph-LLMs in the context of robustness on graph learning, we need to design diverse pipelines to systematically assess the robustness of LLM approaches against adversarial attacks on graphs. Our benchmark evaluation encompasses two pivotal dimensions: LLMs-as-Enhancers and LLMs-as-Predictors. In this section, we will introduce the benchmark design. {Details about the benchmark datasets are provided in Appendix~\ref{sec:sup-dataset}. % Appendices~\ref{sec:sup-threat} and~\ref{sec:sup-dataset}.}

\begin{figure*}[tp]
   \centering
   \begin{minipage}[b]{0.197\textwidth} % 调整 minipage 的宽度
        \centering
       \includegraphics[width=\textwidth]{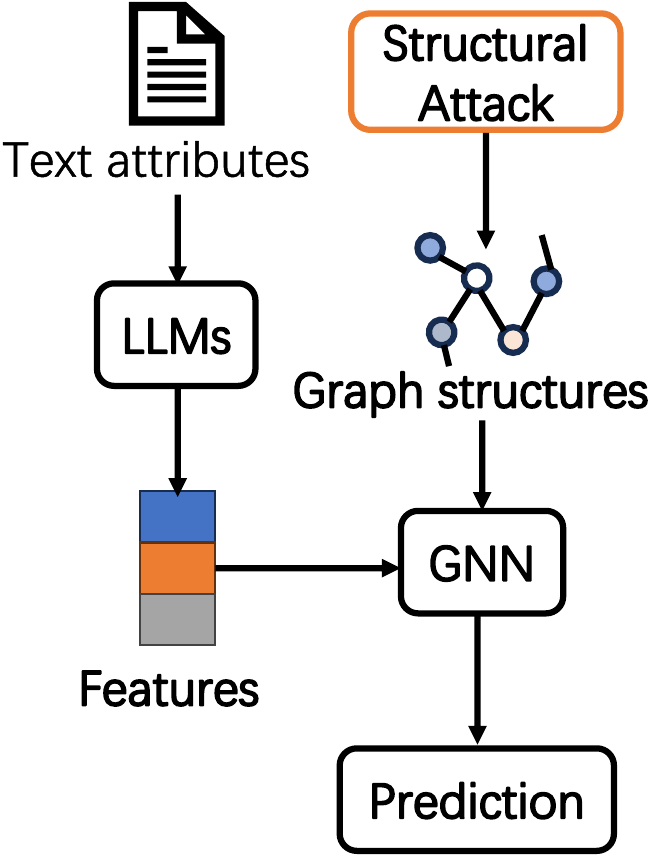}
       \subcaption{}
   \end{minipage}
   \hfill
   \begin{tikzpicture}
       \draw[dashed] (0,0) -- (0,-4); % 添加虚线，调整高度
   \end{tikzpicture}
   \hfill
   \begin{minipage}[b]{0.30\textwidth}
       \centering
       \includegraphics[width=\textwidth]{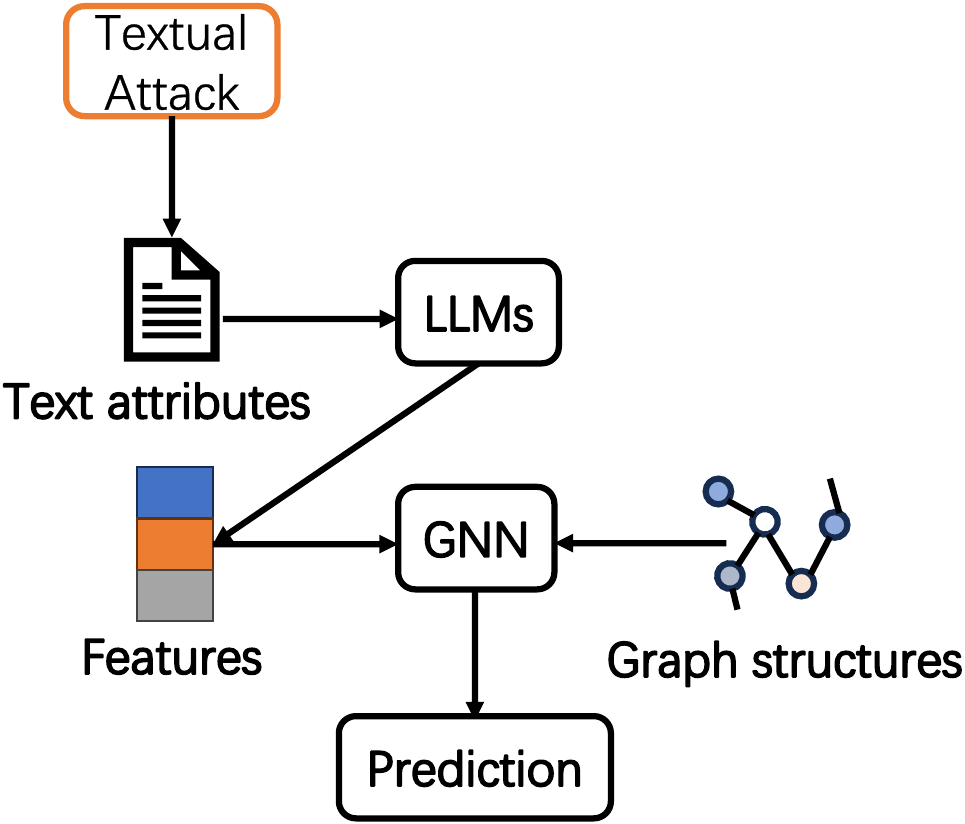}
       \subcaption{}
   \end{minipage}
   \hfill
   \begin{tikzpicture}
       \draw[dashed] (0,0) -- (0,-4); % 添加虚线，调整高度
   \end{tikzpicture}
   \hfill
   \begin{minipage}[b]{0.21\textwidth}
       \centering
       \includegraphics[width=\textwidth]{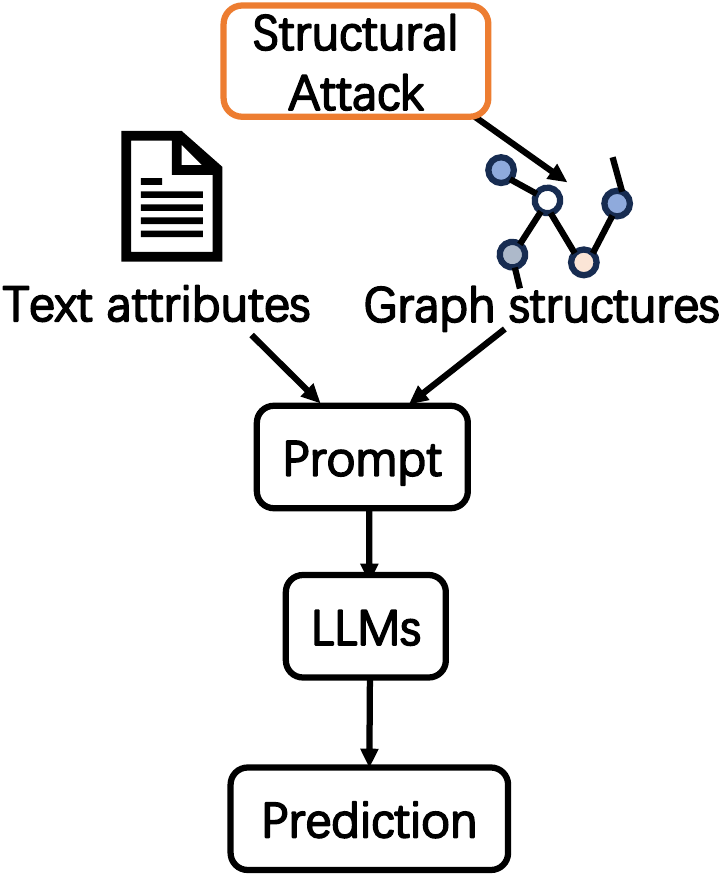}
       \subcaption{}
   \end{minipage}
   \hfill
   \begin{tikzpicture}
       \draw[dashed] (0,0) -- (0,-4); % 添加虚线，调整高度
   \end{tikzpicture}
   \hfill
   \begin{minipage}[b]{0.22\textwidth}
       \centering
       \includegraphics[width=\textwidth]{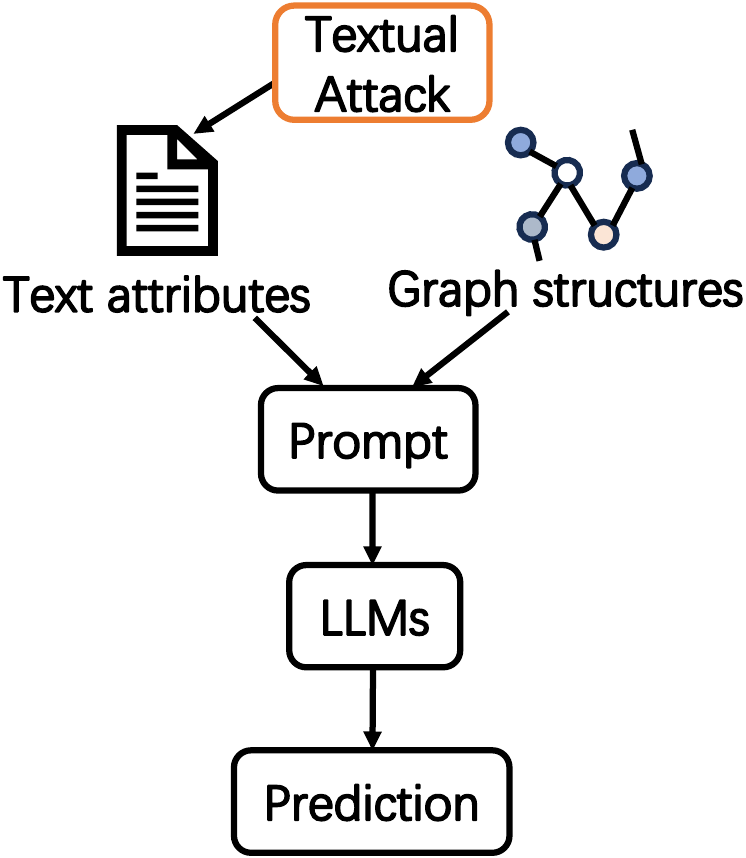}
       \subcaption{}
   \end{minipage}
\caption{Evaluation pipelines: (a)(b) for LLMs-as-Enhancers in structural and textual attacks, respectively; (c)(d) for LLMs-as-Predictors in structural and textual attacks, respectively.}
 \label{fig:pipeline}
 \vskip -1.5em
\end{figure*}
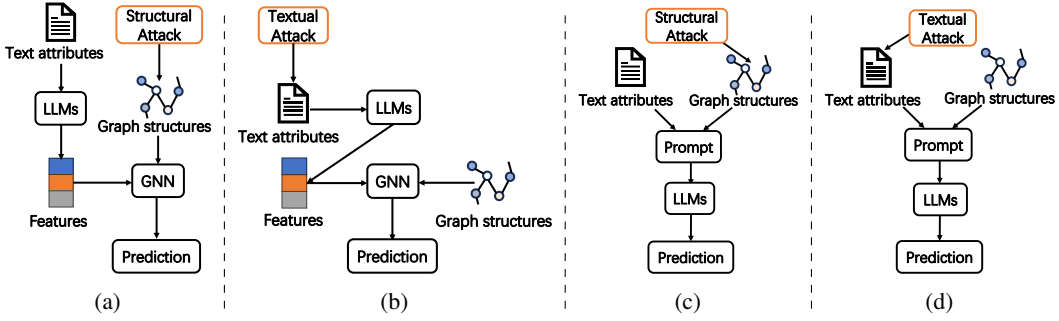

% To delve into the adversarial robustness of Large Language Models (LLMs) on graphs,  we 
% conduct extensive experiments within a unified evaluation framework. Our benchmark assessments encompass two pivotal dimensions: LLMs-as-Enhancers and LLMs-as-Predictors. 

\vspace{-0.5em}
\subsection{Threat Model}
\vspace{-0.5em}

% \label{sec:sup-threat}
We describe the characteristics of the graph adversarial attacks we developed, including both structural and textual attacks, from the following aspects. \textit{(1) Adversary's Goal}: The primary objective is focused on evasion attacks, where the adversary seeks to manipulate the input graph data at inference time to cause the model to make incorrect predictions. In this scenario, the adversaries do not have the authority to change the classifier or its parameters.
\textit{(2) Victim Models:} 
The targets of these attacks include LLMs-as-Predictors and LLMs-as-Enhancers, of which the details will be thoroughly elaborated in Sections~\ref{sec:llm-enhancer} and~\ref{sec:llm-predictors}.  \textit{(3) Adversary’s Knowledge:} {The attacks are designed under white-box and grey-box frameworks, meaning that the adversary either possesses complete knowledge of the model architecture and parameters or not respectively}. White-box attacks are employed during the LLM-as-Enhancers experiments to evaluate robustness in the worst-case scenarios. However, for LLM-as-Predictors, it becomes impractical to perform white-box attacks due to the significant time costs associated with these large, complex models, and thus we adopt a grey-box setting. 
% Thus, we adopt a grey-box setting and utilize \wei{XXX} as the surrogate model to generate grey-box attacks, which will then be transferred to other architectures.

\subsection{LLMs-as-Enhancers}
\label{sec:llm-enhancer}
\vspace{-0.5em}

For LLMs-as-Enhancers, our benchmark provides a fair and comprehensive comparison of existing representative methods from two perspectives: structural attack and textual attack. 

\textbf{Structural attack:}
We have discussed two commonly used methods for structural attacks, PGD~\cite{xu2019topology} and PRBCD~\cite{geisler2021robustness}, in Section~\ref{sec:background}. 
Currently, existing frameworks rely on shallow features such as Bag of Words (BOW) and TF-IDF. In the era of LLMs, it's imperative to examine the impact of LLM features on structural attacks. Therefore, we designed a pipeline for structural attacks using LLMs-as-Enhancers. Specifically, we first generate diverse feature types derived from various LLMs, including SBert, E5, LLaMA, and Angle-LLaMA~\cite{li2023angle}, as well as LLaMA fine-tuned (LLaMA-FT) with LoRa~\cite{hu2021lora}, etc.  We then utilize these features to evaluate the performance of structural attacks. The pipeline is visualized in Figure~\ref{fig:pipeline}(a).

\textbf{Textual attack:} 
% For textual attack, we employed a word-level model called \textbf{SemAttack} \cite{wang2022semattack} to evaluate. Specifically, we randomly sample nodes, initialize perturbations in the word embedding space, and optimize them using gradient descent while maintaining semantic consistency with synonyms from WordNet~\cite{miller1995wordnet}. To enhance efficiency on LLMs, we further modified SemAttack for batch-wise operation.
% Text-attributed graphs may also be vulnerable to text-level attacks. The features required for training Graph Neural Network (GNN) models are derived from these corresponding texts. Consequently, if the texts undergo an attack, it directly impacts the performance of GNN models. 
We conduct an evaluation on text attacks to verify whether LLMs-as-Enhancers can withstand textual attacks compared to traditional text preprocessing techniques. Specifically, in the white-box setting, we first conduct text attacks by using \textbf{SemAttack} \cite{wang2022semattack} on the texts of text-attributed graphs. Then, we encode the texts using different methods such as traditional techniques and LLMs. Finally, we assess their performance on GCN and MLP. Incidentally, to enhance efficiency on LLMs, we modified SemAttack for batch-wise operation instead of word-level processing. The pipeline is illustrated in Figure~\ref{fig:pipeline}(b).

\subsection{LLMs-as-Predictors}
\label{sec:llm-predictors}
\vspace{-0.5em}

For LLMs-as-Predictors, we also perform structural and textual attacks on pre-trained and fine-tuned LLMs, respectively. Different from the attacks on LLMs-as-Enhancers, white-box attacks on the predictors can bring enormous computational costs due to the complexity of the models. Therefore, this study adopts the grey-box setting, choosing LLMs-as-Enhancers as the victim model and transferring the attacked texts or graphs to the LLMs-as-Predictors. For pre-trained models, we utilize the same pipeline in \cite{chen2024exploring}, which describes the graph structure in text and inputs it along with text features directly into GPT-3.5 for prediction.
For fine-tuned models, we perform attacks on InstructGLM \cite{ye2023natural}, which uses LLaMA \cite{touvron2023llama} as the backbone and is fine-tuned on different benchmark datasets.

\textbf{Structural attack:}
Although the LLMs-as-Predictors in this study flatten graph structure into texts and only incorporate a small number of neighbors when predicting a node, it is possible that introducing irrelevant or false neighbors can influence the prediction results. During the structure attack, we use PRBCD as the attack algorithm and choose SBert and GCN as the surrogate model. The perturbed graph then serves as the input of GPT-3.5 or InstructGLM. The pipeline is depicted in Figure~\ref{fig:pipeline}(c).

\textbf{Textual attack:}
The LLMs-as-Predictors directly utilize texts as inputs. Thus, perturbing the input texts is also likely to have impacts to the prediction results. In this study, we first perform SemAttack on the selected nodes with SBert and GCN as the surrogate model. After that, the perturbed texts are used to evaluate GPT-3.5 and InstructGLM. The pipeline is shown in Figure~\ref{fig:pipeline}(d).

\vspace{-0.5em}
\section{Experiments}
\vspace{-0.5em}

\label{sec:experiment}
In this section, we assess the robustness of LLMs against graph adversarial attacks in their two roles: LLMs-as-Enhancers and LLMs-as-Predictors. Specifically, we aim to answer the following questions: \textbf{Q1:} How effective are the LLMs-as-Enhancers on structural attack?  \textbf{Q2:} What is the effectiveness of LLMs-as-Enhancers on textual attacks? 
\textbf{Q3:} How effective are the LLMs-as-Predictors on structural attack? 
\textbf{Q4:} How do LLMs-as-Predictors perform on textual attacks?

% \begin{compactenum}[\textbullet]
%   \item 
% \textbf{Q1:} How effective are the LLMs-as-Enhancers on structural attack? 
% \item
% \textbf{Q2:} What is the effectiveness of LLMs-as-Enhancers in textual attacks? 
% \item
% \textbf{Q3:} How effective are the LLMs-as-Predictors on structural attack? 
% \item
% \textbf{Q4:} How do LLMs-as-Predictors perfsorm in textual attacks?
% \end{compactenum}

% {We provide the experiment setting in Appendix~\ref{sec:sup-setting}.}

%%============================ Enhancer Structural Attack
\subsection{Structural Attack for LLMs-as-Enhancers}
\label{sec:enhancer-struct}

% To answer \textbf{Q1}, we conduct experiments to examine the robustness against structural attack for LLM-as-Enhancers. 

% \usepackage{booktabs}
% \usepackage{graphicx}
\begin{table}[]
\caption{Performance of LLMs-as-Enhancers against \textbf{\underline{0\%, 5\%and 25\% structural attacks}}. \textbf{GAP} represents the percentage (\%) decrease in performance after an attack compared to the clean performance.
N/A indicates that TAPE does not provide the explanation features. We use \textcolor{pink}{pink} to denote the best performance, \textcolor{green}{green} for the second-best, and \textcolor{yellow}{yellow} for the third-best.
}
\resizebox{\textwidth}{!}{%
\setlength{\tabcolsep}{3pt} % added by Wei to reduce the margin between columns
\begin{tabular}{@{}cc|cc|cc|cc|cc|cc|cc|cc|cc|cc@{}}
\toprule
\textbf{} & \textbf{} & \multicolumn{2}{c}{\textbf{BOW}} & \multicolumn{2}{c}{\textbf{TF-IDF}} & \multicolumn{2}{c}{\textbf{SBert}} & \multicolumn{2}{c}{\textbf{E5}} & \multicolumn{2}{c}{\textbf{LLaMA}} & \multicolumn{2}{c}{\textbf{Angle-LLaMA}} & \multicolumn{2}{c}{\textbf{Explanation}} & \multicolumn{2}{c}{\textbf{Ensemble}} & \multicolumn{2}{c}{\textbf{LLaMA-FT}} \\ \midrule
{\textbf{Dataset}} & \textbf{Ptb.} & \textbf{ACC} & \textbf{GAP} & \textbf{ACC} & \textbf{GAP} & \textbf{ACC} & \textbf{GAP} & \textbf{ACC} & \textbf{GAP} & \textbf{ACC} & \textbf{GAP} & \textbf{ACC} & \textbf{GAP} & \textbf{ACC} & \textbf{GAP} & \textbf{ACC} & \textbf{GAP} & \textbf{ACC} & \textbf{GAP} \\ \midrule
\multirow{3}{*}{Pudmed} & 0\% & 74.69 &0\%& 76.86 &0\%& 78.71 &0\%& \cellcolor{yellow}81.83 &0\%& 77.65 &0\%& 75.15 &0\%& \cellcolor{pink}88.84 &0\%& \cellcolor{green}83.59 &0\%& 77.4 &0\%\\
 & 5\% & 72.40 & 15.90\% & 74.68 & 2.8\% & 74.68 & 5.1\% & 79.39 & 3.0\% & 76.68 & 1.3\% & 74.65 & \cellcolor{pink}0.67\% & 87.32 & 1.5\% & 82.55 & \cellcolor{yellow}1.2\% & 76.55 & \cellcolor{green}1.1\% \\
 & 25\% & 62.83 & 24.30\% & 66.18 & 13.9\% & 69.24 & 12.0\% & 71.06 & 13.1\% & 74.62 & \cellcolor{yellow}3.9\% & 72.72 & \cellcolor{green}3.2\% & 82.11 & 7.6\% & 77.85 & 6.9\% & 75.63 & \cellcolor{pink}2.3\% \\
\midrule
\multirow{3}{*}{Arxiv} & 0\% & 50.99 &0\%& 48.39 &0\%& 52.51 &0\%& 57.04 &0\%& \cellcolor{yellow}58.04 &0\%& \cellcolor{green}58.53 &0\%& 54.37 &0\%& \cellcolor{pink}59.9 &0\%& 52.46 &0\%\\
 & 5\% & 42.35 & 16.9\% & 43.32 & \cellcolor{yellow}10.5\% & 47.59 & \cellcolor{green}9.4\% & 48.15 & 15.6\% & 51.76 & 10.8\% & 51.46 & 12.1\% & 47.19 & 13.2\% & 54.33 & \cellcolor{pink}9.3\% & 47.59 & \cellcolor{pink}9.3\% \\
 & 25\% & 18.95 & 62.9\% & 24.36 & 49.7\% & 29.24 & \cellcolor{pink}44.3\% & 25.24 & 55.8\% & 31.67 & 45.4\% & 31.73 & 45.8\% & 21.61 & 60.3\% & 32.82 & \cellcolor{yellow}45.2\% & 29.19 & \cellcolor{green}44.4\% \\\midrule
\multirow{3}{*}{Cora} & 0\% & 78.49 &0\%& 81.46 &0\%& 81.99 &0\%& \cellcolor{pink}83.17 &0\%& 78.13 &0\%& 80.28 &0\%& \cellcolor{green}82.79 &0\%& \cellcolor{yellow}82.57 &0\%& 78.54 &0\%\\
 & 5\% & 73.71 & 3.1\% & 77.23 & 5.2\% & 79.18 & 3.4\% & 80.43 & 3.3\% & 70.91 & 9.2\% & 80.23 & \cellcolor{green}0.06\% & 80.47 & 2.8\% & 80.67 & \cellcolor{yellow}2.3\% & 78.51 & \cellcolor{pink}0\% \\
 & 25\% & 60.72 & 22.6\% & 67.53 & 17.1\% & 72.14 & 12.0\% & 69.84 & 16.0\% & 69.99 & 10.4\% & 76.75 & \cellcolor{pink}4.4\% & 71.32 & 13.9\% & 74.89 & \cellcolor{green}9.3\% & 70.72 & \cellcolor{yellow}10.0\% \\\midrule
\multirow{3}{*}{WikiCS} & 0\% & 74.92 &0\%& 75.96 &0\%& 75.89 &0\%& \cellcolor{yellow}76.71 &0\%& \cellcolor{pink}79.72 &0\%& 73.33 &0\%& N/A & N/A & N/A & N/A & \cellcolor{green}79.69 &0\%\\
 & 5\% & 61.46 & 18.0\% & 62.03 & 18.3\% & 64.44 & 15.1\% & 63.01 & 17.9\% & 70.91 & \cellcolor{green}11.1\% & 67.0 & \cellcolor{pink}8.6\% & N/A & N/A & N/A & N/A & 70.11 & \cellcolor{yellow}12.0\% \\
 & 25\% & 45.34 & 39.5\% & 45.74 & 39.8\% & 50.19 & 33.9\% & 46.38 & 39.6\% & 59.23 & \cellcolor{yellow}25.7\% & 56.75 & \cellcolor{pink}22.6\% & N/A & N/A & N/A & N/A & 60.13 & \cellcolor{green}24.6\% \\\midrule
\multirow{3}{*}{History} & 0\% & 51.25 &0\%& 58.38 &0\%& 65.59 &0\%& \cellcolor{pink}66.96 &0\%& \cellcolor{green}66.73 &0\%& \cellcolor{yellow}66.52 &0\%& N/A & N/A & N/A & N/A & 64.53 &0\%\\
 & 5\% & 49.04 & 4.3\% & 57.57 & \cellcolor{green}1.4\% & 65.6 & \cellcolor{pink}0\% & 64.69 & 3.4\% & 61.64 & 7.6\% & 63.47 & 4.6\% & N/A & N/A & N/A & N/A & 62.99 & \cellcolor{yellow}2.4\% \\
 & 25\% & 38.93 & 24.0\% & 22.66 & 61.2\% & 56.34 & \cellcolor{pink}14.1\% & 55.37 & \cellcolor{green}17.3\% & 54.54 & 18.3\% & 52.75 & 20.7\% & N/A & N/A & N/A & N/A & 53.34 & \cellcolor{yellow}17.4\% \\\midrule
\multirow{3}{*}{Citeseer} & 0\% & 70.74 &0\%& \cellcolor{yellow}73.02 &0\%& \cellcolor{green}74.94 &0\%& \cellcolor{pink}75.14 &0\%& 67.48 &0\%& 71.71 &0\%& N/A & N/A & N/A & N/A & 69.7 &0\%\\
 & 5\% &68.84 & \cellcolor{yellow}2.7\% & 71.40 & \cellcolor{green}2.2\% & 72.60 & 3.1\% & 72.29 & 3.8\% & 70.82 & 5.0\% & 68.46 & 4.5\% & N/A & N/A & N/A & N/A & 68.62 & \cellcolor{pink}1.6\% \\
 & 25\% & 61.76 & 12.7\% & 64.36 & 11.9\% & 64.51 & 13.9\% & 64.38 & 14.3\% & 64.69 & \cellcolor{pink}4.1\% & 68.61 & \cellcolor{green}4.3\% & N/A & N/A & N/A & N/A & 66.17 & \cellcolor{yellow}5.1\% \\\bottomrule
\end{tabular}%
}
\label{tab:enhancer-structure5}
\vskip -1.em
\end{table}

\textbf{Experiment Design.}
To tackle the research question \textbf{Q1}, we enhance text attributes using LLMs and generate new features. These enriched features are then used to train a GCN as the victim model. Specifically, we use PGD to conduct white-box evasion attacks on the structures of small graphs such as Cora, Citeseer, Pubmed, and Wikics, while we employ PRBCD to conduct white-box evasion attacks on the structures of large graphs Arxiv and History. 
% \jt{I think we report both the accuracy denoted as "ACC" and the performance degradation after attacks denoted as "GAP", please rewrite below: }
We vary the perturbation rates at 0\% (clean graphs), 5\%, and 25\%, which represent the ratio of perturbed edges to original edges. Subsequently, we feed the perturbed graphs into different LLMs-as-Enhancers architectures and compare the LLM features with shallow features. To quantify model robustness, we report the test accuracy (ACC) and the percentage accuracy degradation after attacks compared to the original accuracy (GAP).

% \kai{Specifically, we present the accuracy on clean graphs and the performance after an attack. We use the metric 
% $
% \text{GAP} = \frac{\text{clean acc} - \text{attack acc}}{\text{clean acc}}\times100\%
% $ 
% to measure robustness against structural attacks.}

% \textbf{Most LLMs perform well on clean graphs.}

\textbf{Results.} The results are reported in Table~\ref{tab:enhancer-structure5}. Further details, including standard deviations, can be found in Appendix~\ref{sec:sup-enhancer4structure}. From these results, we have the following observations. 

\textit{Performance comparison on clean graphs.}
The features generated by pre-trained language models exhibit better performance on most clean datasets. For instance, SBert and e5-large (E5) show improvements of 2.4\% and 6.5\% respectively on clean Pubmed datasets compared to TF-IDF.

% \textbf{Most LLMs demonstrate robustness against structural attacks.}
\textit{Robustness against structural attacks.}
For 25\% evasion attacks, almost all language models exhibit greater robustness compared to traditional BOW and TF-IDF approaches. For example, in the case of a 25\% evasion attack on Pubmed, while TF-IDF experiences a decrease of 13.9\%, LLaMA only drops by 3.9\%. Moreover, we observe that fine-tuning enhances the robustness of LLMs, as demonstrated by the fact that fine-tuned LLaMA exhibits greater robustness compared to its unfine-tuned counterpart.
Further, by comparing the results on 5\% and 25\% attacks, we find that LLMs-as-Enhancers are more helpful at higher perturbation rates. For example, with a 5\% perturbation budget on Cora, only 4 language models show a lower GAP compared to the shallow BOW features, while this number increases to 7 at the perturbation rate of 25\%.
{However, by examining the accuracy on clean graphs, we note that the performance of LLM features still declines considerably on certain datasets, such as Arxiv, under high perturbation rates, indicating that graph LLMs remain vulnerable to attacks.}

% \textbf{The higher the attack rate, the stronger the robustness of LLMs.}
% \textit{Comparison with different attack rates.}
% \jt{I think we still need to compare with $0\%$ to show graphLLMs are still vulnerable to attacks??}
% We conduct evasion attacks with two different perturbation budgets, 5\% and 25\%, respectively. From these results, we find that LLMs-as-Enhancers are more helpful at higher perturbation rates. For example, with a 5\% perturbation budget on Cora, only 4 language models show a lower GAP compared to the shallow BOW features. With a 25\% perturbation budget, this number increases to 7.

% \textbf{The explanation features generated by GPT also exhibit strong robustness.}
\textit{Robustness of explanation features.}
The explanation features generated by TAPE not only exhibit higher accuracy on clear graphs but also demonstrate good robustness against evasion attacks. When ensemble with LLaMA features, these explanation features show higher accuracy and robustness compared to using LLaMA alone.

\underline{\textbf{Key Takeaways 1:}} Most LLMs demonstrate greater robustness against structural attacks compared to shallow models. 
% \jt{the following is not needed??} {Especially, fine-tuned LLaMA demonstrates even better robustness.}
% \wei{can we make conclusions on the relationship between the power of enhancers and the robustness. e.g., more powerful enhancer exhibits stronger robustness?}

\underline{\textbf{Key Takeaways 2:}} The higher the attack rate, the more robust the features of LLMs compared to shallow features.

\subsection{Textual Attack for LLMs-as-Enhancers}
\label{sec:enhancer-text}

% experiment settings
% To answer \textbf{Q2}, we conduct the white-box evasion attack on LLMs-as-Enhancers, targeting textual attributes. 
% All models are evaluated using the Attack Success Rate (ASR). 

\textbf{Experiment Design.}
To answer \textbf{Q2}, we conduct white-box evasion attacks on LLMs-as-Enhancers, targeting textual attributes.  
%\jt{the following is not needed since ASR is straightforward}as follows: $\text{ASR} = \frac{\mathbf{N_{success}}}{\mathbf{N_{total}}} * 100\%$, where $\mathbf{N_{success}}$ denotes the number of successfully attacked samples. 
We first utilize diverse LLMs-as-Enhancers to transform the text into node embeddings and train a GCN and an MLP. Then we perform the evasion attack at the model inference stage by perturbing text using SemAttack. Then, LLMs are used to transform the perturbed text into node embeddings, which will then be fed into the trained GCN or MLP for inference. For all models and datasets, we randomly sample 200 nodes as target nodes and utilize the Attack Success Rate (ASR)~\cite{wang2022semattack} as the evaluation metric.

% observations
\textbf{Results.}  The results are reported in Table~\ref{tab:enhancer-textual}. Additional details, including standard deviations of performance, are provided in Appendix~\ref{sec:sup-enhancer4text}. From these results, we can draw the following observations. 

\textit{LLaMA performs well against textual attack when MLP is used as the victim model.} 
When using MLP as the victim model, E5 and LLaMA demonstrate greater resilience against SemAttack, with a noticeable downward trend in ASR for SBert, E5, and LLaMA models. For example, on the WikiCS dataset, SBert has an ASR of 62.45\%, while the performance of LLaMA dropped to 22.17\%. Another interesting observation is that BOW shows better robustness than SBert on Cora, Pubmed, and Arxiv. Given that BOW has a limited input of words, the robustness of BOW is likely to come from filtering the perturbed words, whose frequencies are often low.

\textit{For textual attack, GCN as the victim model is more robust compared to MLP as the victim model.} 
In terms of GCN as the victim model, fine-tuned LLaMA achieves the lowest ASR among all datasets, ranging from 2.98\% to 6.91\%. Also, compared to MLP, the ASR for GCN decreases significantly and remains below 20\% across all datasets. 

\underline{\textbf{Key Takeaways 3:}}
Among all models and settings, the fine-tuned model, {LLaMA-FT}, generally exhibits the best robustness against the textual attack on most datasets.

\underline{\textbf{Key Takeaways 4:}}
In the LLMs-as-Enhancers framework, GCN greatly improves the robustness against textual attacks compared to MLP as the victim model.

\begin{table}[]
\centering
\caption{{Performance of LLMs-as-Enhancers against the textual attack. Bold numbers represent the lowest Attack Success Rate (ASR), indicating superior robustness.}}
\label{tab:enhancer-textual}
\setlength{\tabcolsep}{3pt}
\scalebox{0.9}{%
% \resizebox{1\textwidth}{!}{%
\begin{tabular}{@{}c|cc|cc|cc|cc|cc|cc@{}}
\toprule
 & \multicolumn{2}{c|}{Cora} & \multicolumn{2}{c|}{Pubmed} & \multicolumn{2}{c|}{Arxiv} & \multicolumn{2}{c|}{Wikics} & \multicolumn{2}{c|}{History} & \multicolumn{2}{c}{Citeseer} \\ \hline
Features               & MLP         & GCN         & MLP         & GCN         & MLP         & GCN   & MLP         & GCN        & MLP         & GCN         & MLP         & GCN      \\
\hline
BOW               & 62.11  & 9.27   & 41.00  & 8.58  & 72.69   & 15.19 & 67.85  & 3.82   & 74.53  & 18.40  & 68.80  & 18.98 \\ \hline
SBert             & 73.18  & 14.76  & 45.36  & 9.32  & 82.69  & 11.17 & 62.12  & 9.08  & 76.73   & 13.41  & 66.33  & 16.93   \\
E5                & 65.29  & 10.53  & 35.62  & 8.76  & 81.92   & 15.22 & 65.99  & 6.64   & 61.51  & \textbf{6.54}   & 57.10  & 14.69   \\
LLaMA             & 56.49  & 12.58  & 19.66  & 4.95  & 67.87  & \textbf{6.05}  & \textbf{22.17}  & 4.17   & 65.07   & 15.92  & 46.81 & 13.77  \\
LLaMA-FT  & \textbf{40.10} & \textbf{4.37}   & \textbf{16.69}  & \textbf{3.27}  & \textbf{67.71}   & 6.08  & 30.00  & \textbf{2.98}   & \textbf{56.98}  & 6.91   & \textbf{38.46} & \textbf{6.58}  \\    
\bottomrule
\end{tabular}}
\end{table}

% \begin{table}[]
% \centering
% \caption{The robustness of LLMs-as-enhancer against attacks on text features}
% \label{tab:enhancer-textual2}
% \resizebox{1\textwidth}{!}{%
% \begin{tabular}{l|cc|cc|cc}
% \hline
% Dataset & \multicolumn{2}{c}{Wikics} & \multicolumn{2}{c}{History} & \multicolumn{2}{c}{Citeseer} \\ \hline
% ASR               & MLP         & GCN         & MLP          & GCN         & MLP         & GCN         \\
% \hline
% BOW               & 67.85±4.88  & 3.82±1.46   & 74.53±10.13  & 18.40±6.00  & 68.80±6.99  & 18.98±4.67   \\
% SBert             & 62.12±4.32  & 9.08±2.34   & 76.73±8.79   & 13.41±3.86  & 66.33±6.09  & 16.93±3.31    \\
% E5                & 65.99±5.05  & 6.64±2.47   & 61.51±15.43  & 6.54±1.04  & 57.10±6.25  & 14.69±3.73    \\
% LLaMA             & 22.17±4.96  & 4.17±0.80   & 65.07±9.50   & 15.92±8.47  & 46.81±14.07 & 13.77±5.34    \\
% Fine-tuned LLaMA  & 30.00±2.07  & 2.98±1.83   & 56.98±14.55  & 6.91±3.64   & 38.46±11.19 & 6.58±2.97    \\     
% \hline
% \end{tabular}}
% \end{table}

%%============================ Predictor Structural Attack
\subsection{Structual Attack for LLMs-as-Predictors}
\label{sec:predictor-struct}

\textbf{Experiment Design.} To answer \textbf{Q3}, we explore the robustness of LLM-as-Predictors against graph structural attacks, using GPT-3.5 and InstructGLM as the selected predictors.  Given the difficulty in directly attacking these predictors due to their large number of parameters or lack of model access, we adopt a grey-box setting. In this setting, LLMs-as-Enhancers are used as surrogate models during adversarial attacks, and the resulting attacked graph structures are then fed into the LLMs-as-Predictors. 
When employing GPT-3.5 in the graph domain, we follow the approach in \cite{chen2024exploring} and evaluate 200 nodes in the test set. For InstructGLM, we directly attack the pre-processed datasets provided by the author.

\textbf{Results.} 
The results of InstrucGLM are reported in Table~\ref{tab:predictor-structural2}. More results about GPT-3.5 are presented in Figure~\ref{fig:predictor-structual} in Appendix~\ref{sec:sup-predictor4structure}. 
Based on these results, we can make the following observations.

\textit{GPT-3.5 shows the strongest robustness against structural attack in the zero-shot setting.}  
% We evaluate the model under various few-shot and zero-shot settings with the summary of two-hop neighbors\wei{what does this mean?}.

We evaluate the model under various few-shot and zero-shot settings with the input of summarized two-hop neighbors, as illustrated in~\cite{chen2024exploring}.
The results demonstrate that GPT-3.5 maintains the highest robustness in the zero-shot setting, with minimal performance degradation even under significant perturbations. We conjecture that the neighbor sampling and summarizing processes likely mitigate noise introduced by structural alterations. Although GAT also uses the migrated attacked graph structure from GCN, its accuracy drops much faster than that of GPT-3.5.

% \textit{InstructGLM shows the strongest robustness against structural attack in the zero-shot setting.} 
\textit{Similarly, InstructGLM also shows stronger robustness against structural attack compared to GCN.}
While GCN experiences a noticeable accuracy decrease after 5\% structural perturbation, InstructGLM maintains performance close to that on the clean graph. Like GPT-3.5, InstructGLM also employs neighbor sampling and its robustness may benefit from this procedure.
% While GCN suffers an obvious decrease in accuracy after 5\% structural perturbation, InstructGLM remains a performance close to the result on the clean graph. Like GPT-3.5, InstructGLM also adopts neighbor sampling.

\underline{\textbf{Key Takeaways 5:}} LLMs-as-Predictors show strong robustness against the structural attack, especially in the zero-shot setting.

\begin{table}[tp]

\centering
\caption{The robustness of the predictor InstructGLM against \underline{\textbf{5\% structural attack}}. \textbf{GAP} refers to the percentage (\%) performance decrease after an attack compared to the clean ACC. The bold font is used to highlight the smallest gap.}
\label{tab:predictor-structural2}
\setlength{\tabcolsep}{3pt}
\resizebox{0.9\textwidth}{!}{%
\begin{tabular}{l|ccc|ccc|ccc}
\toprule
Dataset    & \multicolumn{3}{c|}{Cora} & \multicolumn{3}{c|}{Pubmed} & \multicolumn{3}{c}{Arxiv} \\
\midrule
model \textbackslash{} Perturbation Rate  & Clean  & Attack  & GAP   & Clean  & Attack  & GAP   & Clean  & Attack  & GAP     \\
\midrule
GCN                           & 87.45   & 81.73  & 6.54\%         & 87.07  & 84.25 & 3.24\%       & 57.5  & 52.94  & 7.93\%           \\
InstructGLM (structure-aware) & 82.47   & 80.73  & \textbf{2.11\%}         & 91.63  & 91.10 & \textbf{0.58\%}       & 72.87 & 71.84  & \textbf{1.41\%}       \\ 
\bottomrule
\end{tabular}}
\vskip -3 em
\end{table}

% MLP                         & 67.53        & \textbackslash{}       & 86.21         & \textbackslash{}        & 5.86             & \textbackslash{}           \\
% InstructGLM（structure-free）  & 75.65        & \textbackslash{}       & 92.34         & \textbackslash{}        & 75.65            & \textbackslash{}           \\

%%============================ Predictor Textual Attack
\subsection{Textual Attack for LLMs-as-Predictors}
\label{sec:predictor-text}

\textbf{Experiment Design.} 
To answer \textbf{Q4}, we explore the performance of GPT-3.5 and InstructGLM with perturbed texts as inputs. The perturbed texts are generated by SemAttack and used as adversarial inputs for GPT-3.5 and InstructGLM to evaluate the robustness of LLMs-as-Predictors against textual attacks. Following the experiment design in \textbf{Q3}, the experiment is conducted in a grey-box setting. We use GCN with SBert embeddings as the surrogate model, randomly sampling 200 target nodes for attack and evaluation.

% \textbf{GPT-3.5 against Textual Attack.} As shown in Table \ref{tab:predictor-textual}, our experiment reveals that GPT-3.5 has strong robustness compared to MLP but fails to exceed GCN and GAT in all few shot settings. On the other hand, GAT exhibits the strongest robustness, followed by GCN. However, on Cora and Pubmed, the Attack Success rate of GPT-3.5 is close to GCN, ranging from 0.55\% to 1.70\%. 

% \textbf{InstructGLM against Textual Attack.}
% Different from the observations above, the fine-tuned InstructGLM shows strong robustness on Cora and Pubmed datasets. While there is an Attack Success rate of 30.37\% for GCN on Cora, InstructGLM is able to resist the textual perturbation and achieves an asr of 1.21\% with structural information incorporated. However, we also observe that InstructGLM somehow becomes vulnerable on the Arxiv dataset, with an asr of up to 65.28\%.

\textbf{Results.}  The results of InstructGLM are presented in Table~\ref{tab: fine_pred_text}. More results about GPT-3.5 are presented in Table~\ref{tab:sup-predictor-textual} in Appendix~\ref{sec:sup-predictor4text}.
These results lead us to the following observations.

\textit{GPT-3.5 shows stronger robustness compared to MLP but failed to exceed GCN.}
As shown in Table~\ref{tab:sup-predictor-textual}, our experiment reveals that GPT-3.5 has strong robustness compared to MLP but fails to exceed GCN and GAT in all few shot settings. On the other hand, GAT exhibits the strongest robustness, followed by GCN. However, on Cora and Pubmed, the Attack Success rate of GPT-3.5 is close to GCN, ranging from 0.55\% to 1.70\%.

\textit{InstructGLM shows the strongest robustness compared to both MLP and GCN.}
Different from the above observations of GPT-3.5, the fine-tuned InstructGLM shows stronger robustness on the three datasets compared to GCN. While there is an ASR of 30.37\% for GCN on Cora, InstructGLM can resist the textual perturbation and achieves an ASR of 1.21\% with structural information incorporated. The 0\% ASR result of MLP on the Arxiv dataset is due to the low ACC, as we can not find a sample that is both predicted correctly before the attack and predicted wrong after the attack. Also, this low accuracy is likely to be caused by the dataset used by InstructGLM, which only utilizes titles of a few words as the node attributes.

% However, we also observe that InstructGLM somehow becomes vulnerable on the Arxiv dataset, with an ASR of up to 65.28\%. 
% \wei{any more analysis of this observation? e.g., we conjecture that xxxx.}

\underline{\textbf{Key Takeaways 6:}}
LLMs-as-Predictors exhibit stronger robustness against textual attacks compared to MLP. However, GPT-3.5, which is not fine-tuned, shows poorer robustness compared to GCN.

\begin{table*}[tp]
\centering
\caption{The robustness of the predictor InstructGLM against textual attacks. The bold font is used to highlight the lowest Attack Success Rate (ASR).}
\label{tab: fine_pred_text}
\resizebox{0.8\textwidth}{!}{%
\begin{tabular}{l|cc|cc|cc}
\toprule
Dataset  & \multicolumn{2}{c|}{Cora} & \multicolumn{2}{c|}{Pubmed} & \multicolumn{2}{c}{Arxiv} \\ \hline
Model     & \multicolumn{1}{c|}{Clean ACC}   & \multicolumn{1}{c|}{ASR}        & \multicolumn{1}{c|}{Clean ACC}   & \multicolumn{1}{c|}{ASR}    & \multicolumn{1}{c|}{Clean ACC}   & \multicolumn{1}{c}{ASR}     \\ \midrule
MLP   & \multicolumn{1}{c|}{67.53}       & \multicolumn{1}{c|}{30.37}      & \multicolumn{1}{c|}{86.92}        &  \multicolumn{1}{c|}{10.11}      & \multicolumn{1}{c|}{5.86}        &   \multicolumn{1}{c}{\textbf{0}}     \\ 
GCN   & \multicolumn{1}{c|}{88.01}       & \multicolumn{1}{c|}{10.22}      & \multicolumn{1}{c|}{86.82}        &  \multicolumn{1}{c|}{3.55}      & \multicolumn{1}{c|}{58.85}       &  \multicolumn{1}{c}{6.61}      \\ 
% InstructGLM (structure-free) & \multicolumn{1}{c|}{76.00}       &  \multicolumn{1}{c|}{2.63}     & \multicolumn{1}{c|}{91.50}       &  \multicolumn{1}{c|}{\textbf{0}}      & \multicolumn{1}{c|}{73.00}       &  \multicolumn{1}{c}{25.34}   \\   
InstructGLM (structure-aware) & \multicolumn{1}{c|}{82.50}       &  \multicolumn{1}{c|}{\textbf{1.21}}     & \multicolumn{1}{c|}{91.50}        &  \multicolumn{1}{c|}{\textbf{1.09}}      & \multicolumn{1}{c|}{74.87}      &  \multicolumn{1}{c}{3.42}   \\ 
\bottomrule
\end{tabular}}
\end{table*}

\subsection{Analysis for Structural Attack}
% Now we analyze how large models can enhance robustness against adversarial attacks. To explore the reasons behind and answer \textbf{Q5: Why do LLMs improve robustness to structural attacks?}, we perform analysis from the following perspectives:

Since Sections \ref{sec:enhancer-struct} and \ref{sec:predictor-struct} have shown the robustness of Graph-LLMs against the structure attack,  we conduct analysis from the following perspectives to explore the reasons behind such robustness.

\begin{figure*}[tp]
   \centering
   \begin{minipage}[b]{0.18\textwidth}
        \centering
       \includegraphics[width=\textwidth]{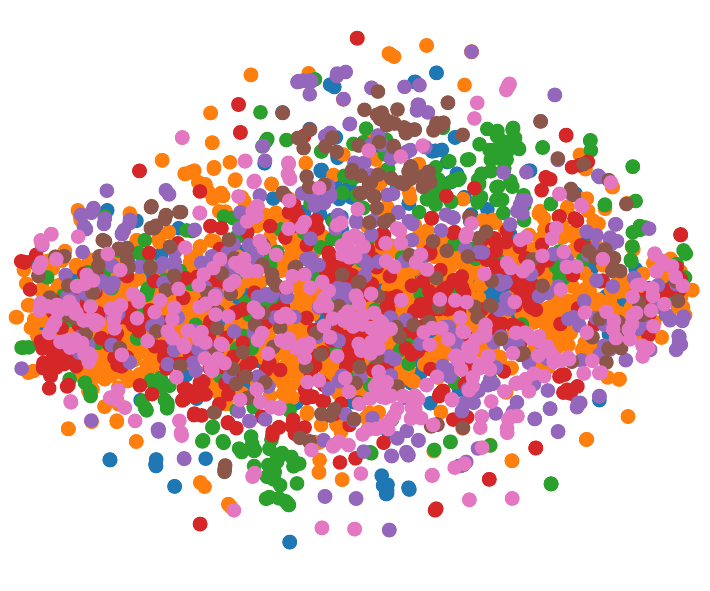}
       \subcaption{BOW}%\usepackage{subcaption}
   \end{minipage}
   \hfill
   \begin{minipage}[b]{0.18\textwidth}
       \includegraphics[width=\textwidth]{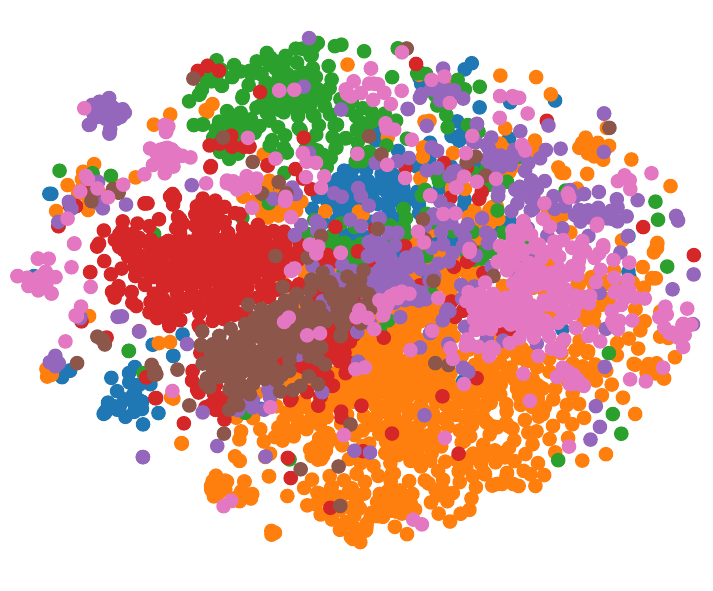}
       \subcaption{TF-IDF}
   \end{minipage}
   \hfill
   \begin{minipage}[b]{0.18\textwidth}
       \includegraphics[width=\textwidth]{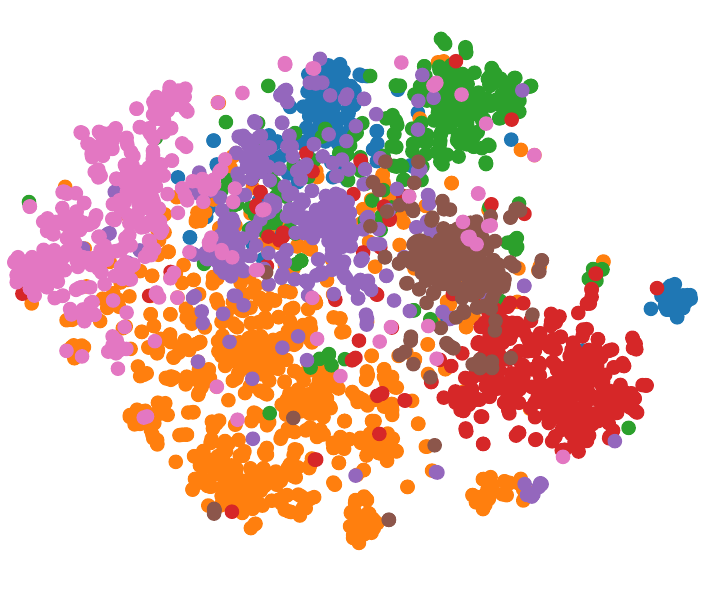}
       \subcaption{SBert}
   \end{minipage}
   \hfill
   \begin{minipage}[b]{0.18\textwidth}
       \includegraphics[width=\textwidth]{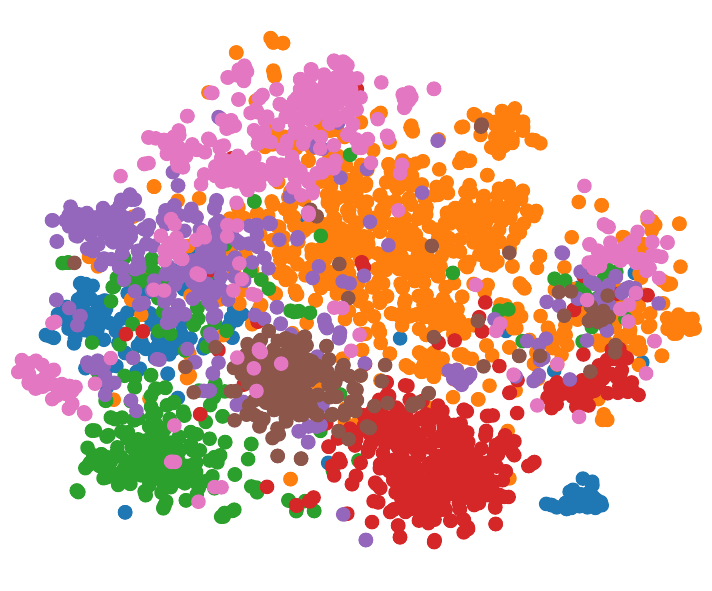}
       \subcaption{E5}
   \end{minipage}
   \hfill
   \begin{minipage}[b]{0.18\textwidth}
       \includegraphics[width=\textwidth]{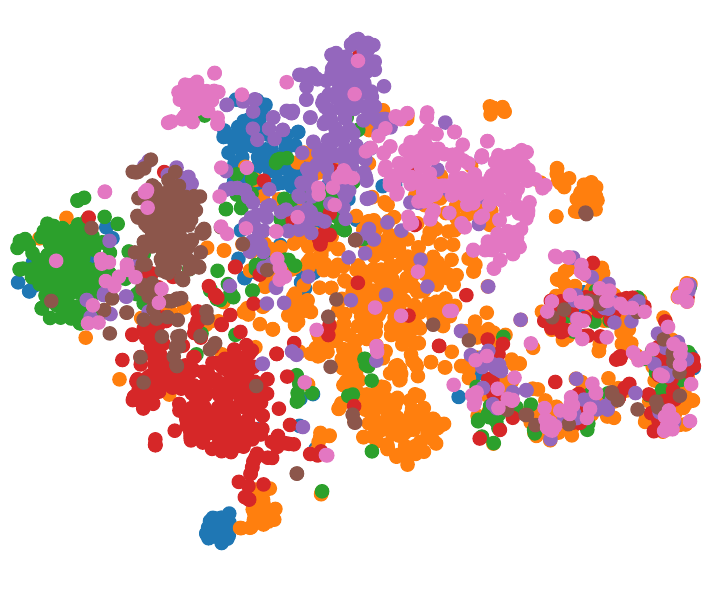}
       \subcaption{LLaMA}
   \end{minipage}
   \caption{t-SNE visualization of different initial features of Cora. Different colors refer to different classes.}
   \label{fig:t-sne}
\vskip -1.3 em
\end{figure*}
\textbf{t-SNE visualization.}
First, we examine t-SNE visualizations of various initial features and find that initial features generated by language models are more distinguishable in categories compared to traditional features as shown in Fig.~\ref{fig:t-sne}. 

\textbf{DBI.} To further evaluate the separability of input features, we use the Davies-Bouldin Index (DBI)~\cite{davies1979cluster}, as shown in Table~\ref{tab:DBI}. The DBI score represents the average similarity measure between each cluster and its most similar cluster, with lower scores indicating better clustering quality. A score of zero is ideal, signifying optimal clustering. We find that initial LLM features have lower initial DBI scores. We further examine the DBI of embeddings both before and after the attack, and the differences therein. Notably, embeddings from larger models exhibit a smaller decrease in DBI scores post-attack, as indicated by the DBI Diff in the Table~\ref{tab:DBI}.

% \textbf{Ablation study.}
% In addition, we conduct an ablation study to evaluate the impact of structure and features separately. First, we examine the performance of an MLP without structural information. Then, we replace the original features with the nodes' degrees to observe the changes and get the new performance. Our findings indicate that features play a crucial role in prediction. 
% \begin{table}[tp]
% \centering
% \caption{DBI and homophily of Cora}
% \label{tab:DBI}
% \resizebox{1\textwidth}{!}{%
% \begin{tabular}{l|ccccccccc}
% \hline
%  & ACC & Init DBI↓ &Embed DBI↓ &Post-Attack DBI↓ &DBI Diff& 25\% Attack&GAP & Homophily 0.8137 \\ \hline
% BOW & 78.49 & 9.34 &1.60 &2.75 &1.15 & 60.72&22.68\% & 0.6594 

%  \\ 
% TF-IDF & 81.46 & 8.85 &1.24 &1.89 &0.65 & 67.53&17.10\% & 0.6597 

%  \\ 
% SBert & 81.99 & 4.47 &1.28 &1.56 &0.28 & 72.14&12.00\% & 0.6964 

% \\ 
% E5 & 83.17 & 5.92 &1.27 &1.66 &0.39 & 69.84&16.00\% & 0.6739

%  \\ 
% LLaMA & 78.13 & 4.88 &1.83 &1.90 &0.07 & 69.99&10.41\% & 0.6821 

%  \\ \hline
% \end{tabular}}
% \end{table}
\begin{table}[tp]
\centering
\caption{DBI and homophily of Cora. ``Init DBI'' indicates the DBI of initial features, ``Embed DBI'' represents the DBI of trained embedding, ``Post-Attack DBI'' refers to the DBI of embedding after an attack, ``DBI Diff'' donates the difference between Embed DBI and Post-Attack DBI. The homophily value of the graph before being attacked is 0.81.}
\label{tab:DBI}
\resizebox{1\textwidth}{!}{%
\begin{tabular}{l|ccccccccc}
\hline
 & GCN & MLP &Init DBI↓ &Embed DBI↓ &Post-Attack DBI↓ &DBI Diff↓ &GAP↓ & Homophily (0.81) \\ \hline
BOW & 78.49& 55.90 & 9.34 &1.60 &2.75 &1.15 
&22.68\% & 0.66 

 \\ 
TF-IDF & 81.46 & 65.85& 8.85 &1.24 &1.89 &0.65 
&17.10\% & 0.66 

 \\ 
SBert & 81.99& 70.80 & 4.47 &1.28 &1.56 &0.28 
&12.00\% & 0.70 

\\ 
E5 & 83.17 & 69.16& 5.92 &1.27 &1.66 &0.39 
&16.00\% & 0.67

 \\ 
LLaMA & 78.13& 67.48 & 4.88 &1.83 &1.90 &0.07 
&10.41\% & 0.68 

 \\ \hline
\end{tabular}}
\end{table}
\textbf{Homophily.}
Additionally, we analyze the homophily of the Cora dataset before and after attacks, as shown in Table~\ref{tab:DBI}. It discovers that the pre-attack Cora dataset has a homophily of 0.81, and after the attack, Cora with shallow features exhibits lower homophily. 

Based on the above results, we find that robustness is strongly positively correlated with the quality of features, indicating that higher distinguishability of features leads to stronger robustness and higher homophily after attacks. This could be attributed to the richer information present in features generated by pre-trained language models, resulting in higher distinguishability in clustering. The higher the quality of the features, the less the model depends on the structure. Therefore, high-quality features of LLMs can be more robust against structural attacks.

% Clearer boundaries between clusters make it more difficult to connect the edges of different categories.

\subsection{Analysis for Textual Attack}
From the experiments of the LLMs-as-Enhancers against textual attack in Section~\ref{sec:enhancer-text}, we observe that LLaMA performs much better than other smaller models and the fine-tuned LLaMA exhibits the best robustness. In addition, GCN demonstrates much stronger robustness compared to MLP. To explore the reasons behind this, we perform analysis from feature and structure perspectives. 
\begin{wrapfigure}{R}{0.32\linewidth}
    \vskip -1.8em
    \centering
    \includegraphics[width=\linewidth]{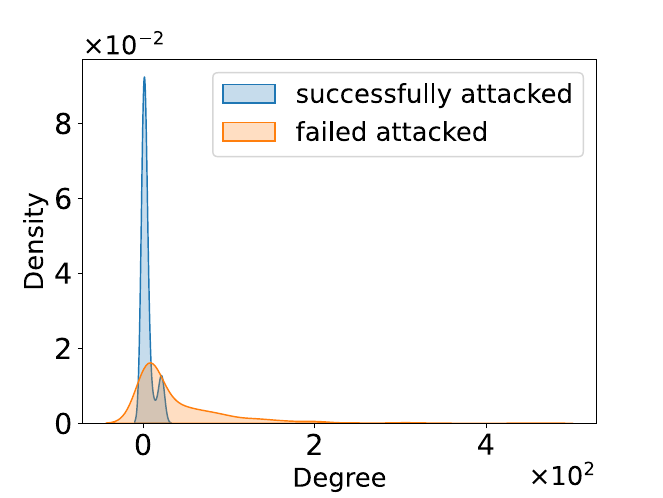}
    \vskip -0.5em
\caption{\footnotesize
Centrality distributions of nodes being attacked successfully and unsuccessfully on WikiCS.}
    \vskip -2.1em
    \label{fig:example}
\end{wrapfigure}

\textbf{Feature Perspective.} From the feature perspective, we assume that text attributes can have an impact on the attack success rate. Specifically, we first explore some basic {indicators} like text entropy, text length, and the number of words among successfully attacked nodes and failed attacked nodes, as shown in Table~\ref{tab:LLaMA_analysis}. Based on the observation with LLaMA and GCN as the victim model, it is obvious that the successfully attacked nodes tend to have smaller entropy (less richness of texts), shorter texts, and smaller amounts of words. By comparing the DBI of the fine-tuned and not fine-tuned model, 
% we also find that the fine-tuned model always has a smaller DBI between the embeddings of successfully attacked nodes and failed attacked nodes.
{we also observe that the fine-tuned model always has a smaller DBI compared to the fine-tuned model}.

\begin{table*}[tp]
\centering
\caption{Comparisons between successfully attacked nodes and failed attacked nodes from the feature perspective (\textbf{success / failed}), with LLaMA and GCN as the victim model.}
\label{tab:LLaMA_analysis}
\resizebox{1\textwidth}{!}{%
\begin{tabular}{l|cc|cc|cc}
\toprule
Dataset       & \multicolumn{2}{c|}{Pubmed}           & \multicolumn{2}{c|}{Citeseer}       & \multicolumn{2}{c}{History}      \\ \midrule
              & \multicolumn{1}{c|}{Fine-tuned}      & \multicolumn{1}{c|}{w/o Fine-tuned}     & \multicolumn{1}{c|}{Fine-tuned}       & \multicolumn{1}{c|}{w/o Fine-tuned}  & \multicolumn{1}{c|}{Fine-tuned}      & \multicolumn{1}{c}{w/o Fine-tuned}    \\ \hline
Entropy       & \multicolumn{1}{c|}{6.25/6.40}       & \multicolumn{1}{c|}{6.24/6.41}          & \multicolumn{1}{c|}{5.90/6.09}        & \multicolumn{1}{c|}{5.86/6.09}       & \multicolumn{1}{c|}{5.45/6.21}       & \multicolumn{1}{c}{5.28/6.25}         \\ 
Text Length   & \multicolumn{1}{c|}{206.10/237.22}   & \multicolumn{1}{c|}{197.94/237.64}      & \multicolumn{1}{c|}{125.17/148.94}    & \multicolumn{1}{c|}{124.07/149.38}   & \multicolumn{1}{c|}{114.71/250.45}   & \multicolumn{1}{c}{95.51/237.71}     \\ 
Words         & \multicolumn{1}{c|}{108.48/122.04}   & \multicolumn{1}{c|}{108.69/122.89}      & \multicolumn{1}{c|}{78.60/90.80}      & \multicolumn{1}{c|}{77.57/90.99}     & \multicolumn{1}{c|}{73.45/139.59}    & \multicolumn{1}{c}{63.79/138.39}     \\ 
DBI           & \multicolumn{1}{c|}{4.77}            & \multicolumn{1}{c|}{4.87}               & \multicolumn{1}{c|}{5.01}             & \multicolumn{1}{c|}{5.24}           & \multicolumn{1}{c|}{4.65}            & \multicolumn{1}{c}{5.02}     \\ 
\bottomrule
\end{tabular}}
\vskip -2. em
\end{table*}

\textbf{Structure Perspective.}
From the structure perspective, we investigate the relations between the degree centrality of nodes and the attack success rate. As shown in Fig.~\ref{fig:example}, we use SBert and GCN as the victim model and visualize the distribution of degrees from successfully attacked nodes and failed attacked nodes respectively. The results clearly show that successfully attacked nodes often have smaller degrees, indicating that \textbf{nodes with less structural information in the graph are more vulnerable to textual attacks}. Additionally, similar patterns are observed for eigenvector centrality and PageRank values, which we detail in Appendix~\ref{sec:sup-analysis} due to page limit.

% \begin{figure}[h]
%   \centering
%   \includegraphics[width=1\columnwidth]{figures/analysis_centrality.pdf} 
%   \caption{Centrality distributions of the node being attacked successfully and unsuccessfully. We use Sentence-Bert as the victim model and gather all attacked results from the Wikics dataset.}
%   \label{fig:analysis_centrality}
% \end{figure}

\section{Conclusion and Future Directions}
\label{sec:future}
This work introduces a comprehensive benchmark for exploring the potential of LLMs in context of adversarial attacks on graphs. Specifically, we investigate the robustness against graph structural and textual attacks in two dimensions: LLMs-as-Enhancers and LLMs-as-Predictors. 
% Extensive experiments reveal that using LLMs to generate features for node attributes enhances robustness against  structural attacks, particularly at high attack rates. Additionally, LLMs show excellent robustness against textual attacks, with GCN proving significantly more robust than MLP as a surrogate model.
Through extensive experiments, we find that, compared to shallow models, both LLMs-as-Enhancers and LLMs-as-Predictors offer superior robustness against structural and textual attacks.
Despite these promising results, several critical challenges and research directions remain worthy of future investigation.

\textbf{Rethinking Textual Attack.} Based on the observations above, we realize that textual attacks can significantly affect the prediction of individual samples. However, when GCN serves as the victim model, the incorporation of neighbor information helps mitigate these perturbations, significantly reducing the attack's effectiveness. From an attack perspective, how the resistance capability of the GCN-based victim models can be weakened deserves attention, especially for attacking stronger Graph-LLMs.

% From an attack perspective, how can the resistance capability of GCN-based victim model be weakened deserves attention.

\textbf{Combining Textual and Structural Attack on Graphs.}
To enhance attack capabilities, a combined framework that perturbs both text attributes and graph structure is needed. However, challenges such as integrating textual and structural attacks to improve attack efficiency remain unsolved. In this study, we provide preliminary results in the Appendix~\ref{sec:sup-combine}. Our experiment shows that adding additional textual perturbations on top of structural perturbations can further degrade model performance.

% {Compared to 5\% perturbation on graph structure, adding an additional 5\% perturbation on node attributes generally decreases the overall accuracy by up to 1.02\%.}

% \textbf{Designing Structural Attack and Defense based on Homophily}
% In essence, existing structural attack methods primarily disrupt the original homophily assumptions. They reduce the predictive capabilities by eliminating homophilous edges or creating heterophilous links. While these attack methods are effective, they can be easily detected by defense mechanisms based on homophily. Therefore, when designing structural attack methods, it is crucial to consider homophily to avoid significant changes with it.

\textbf{Rethinking Graph-LLMs.}
From the results in Table~\ref{tab:enhancer-structure5}, we can conclude that the angle-optimized Angle-LLaMA, which is more suitable for text encoding, exhibits better robustness against adversarial attacks compared to LLaMA. This phenomenon may inspire us to design better Graph-LLMs for text attribute encoding. Finally, we can use LLMs to perform attacks on graphs by generating harmful structures and text attributes.

% \input{sections/6.conclusion}
% \section*{References}

% \bibliographystyle{ACM-Reference-Format}
% \section{Reference}
\bibliographystyle{unsrt}
\bibliography{main}

%%%%%%%%%%%%%%%%%%%%%%%%%%%%%%%%%%%%%%%%%%%%%%%%%%%%%%%%%%%%
\newpage
\section*{Checklist}

% %%% BEGIN INSTRUCTIONS %%%
% The checklist follows the references.  Please
% read the checklist guidelines carefully for information on how to answer these
% questions.  For each question, change the default \answerTODO{} to \answerYes{},
% \answerNo{}, or \answerNA{}.  You are strongly encouraged to include a {\bf
% justification to your answer}, either by referencing the appropriate section of
% your paper or providing a brief inline description.  For example:
% \begin{itemize}
%   \item Did you include the license to the code and datasets? \answerYes{See Section~\ref{gen_inst}.}
%   \item Did you include the license to the code and datasets? \answerNo{The code and the data are proprietary.}
%   \item Did you include the license to the code and datasets? \answerNA{}
% \end{itemize}
% Please do not modify the questions and only use the provided macros for your
% answers.  Note that the Checklist section does not count towards the page
% limit.  In your paper, please delete this instructions block and only keep the
% Checklist section heading above along with the questions/answers below.
% %%% END INSTRUCTIONS %%%

\begin{enumerate}

\item For all authors...
\begin{enumerate}
  \item Do the main claims made in the abstract and introduction accurately reflect the paper's contributions and scope?
    \answerYes{}
  \item Did you describe the limitations of your work?
    \answerYes{} See Section~\ref{sec:sup-impact}
  \item Did you discuss any potential negative societal impacts of your work?
    \answerYes{} See Section~\ref{sec:sup-impact}
  \item Have you read the ethics review guidelines and ensured that your paper conforms to them?
    \answerYes{}
\end{enumerate}

\item If you are including theoretical results...
\begin{enumerate}
  \item Did you state the full set of assumptions of all theoretical results?
    \answerNA{}
	\item Did you include complete proofs of all theoretical results?
    \answerNA{}
\end{enumerate}

\item If you ran experiments (e.g. for benchmarks)...
\begin{enumerate}
  \item Did you include the code, data, and instructions needed to reproduce the main experimental results (either in the supplemental material or as a URL)?
    \answerYes{}
  \item Did you specify all the training details (e.g., data splits, hyperparameters, how they were chosen)?
    \answerYes{}
	\item Did you report error bars (e.g., with respect to the random seed after running experiments multiple times)?
    \answerYes{}
	\item Did you include the total amount of compute and the type of resources used (e.g., type of GPUs, internal cluster, or cloud provider)?
    \answerYes{} See Section~\ref{sec:sup-setting}
\end{enumerate}

\item If you are using existing assets (e.g., code, data, models) or curating/releasing new assets...
\begin{enumerate}
  \item If your work uses existing assets, did you cite the creators?
    \answerYes{}
  \item Did you mention the license of the assets?
    \answerYes{} See Section~\ref{sec:sup-dataset}
  \item Did you include any new assets either in the supplemental material or as a URL?
    \answerYes{}
  \item Did you discuss whether and how consent was obtained from people whose data you're using/curating?
    \answerNA{}
  \item Did you discuss whether the data you are using/curating contains personally identifiable information or offensive content?
    \answerNA{}
\end{enumerate}

\item If you used crowdsourcing or conducted research with human subjects...
\begin{enumerate}
  \item Did you include the full text of instructions given to participants and screenshots, if applicable?
    \answerNA{}
  \item Did you describe any potential participant risks, with links to Institutional Review Board (IRB) approvals, if applicable?
    \answerNA{}
  \item Did you include the estimated hourly wage paid to participants and the total amount spent on participant compensation?
    \answerNA{}
\end{enumerate}

\end{enumerate}

\newpage
\appendix
% \section{Appendix}

\section{Formulations and Background}
\label{sec:sup-background}
We begin by providing preliminaries on graph neural networks, and then formalize the graph adversarial attacks. Finally, we introduce the developments in large language models on graphs.
% \subsection{Graph Neural Networks}

\textbf{Notations.} We define a graph as $G = (V,E)$, where $V$ denotes the node set and $E$ represents the edge set. We employ $\mathbf{X} \in \mathbb{R}^{N \times d}$ to denote the node feature matrix, where $N$ is the number of nodes and 
$d$ is the dimension of the node features.  Furthermore, we use the matrix $\mathbf{A} \in \mathbb{R}^{N \times N}$ to signify the adjacency matrix of $G$. 
% In this matrix, each $\mathbf{A}_{ij}$ = 1 denotes a connection between nodes $v_i$ and $v_j$ in $G$. 
Finally, the graph data can be denoted as $G = (\mathbf{A},\mathbf{X})$. 

\textbf{Graph Neural Networks.} GCN~\cite{kipf} is one of the most representative models of GNNs, utilizing aggregation and transformation operations to model graph data. Unlike GCN, which treats all neighbors equally, GAT~\cite{velivckovic2017graph} assigns different weights to different nodes within a neighborhood during aggregation.

% Based on these notations, let's define a two-layer graph convolutional network (GCN) with parameters $\theta = (\mathbf{W}^{(1)}, \mathbf{W}^{(2)})$. The function $f_{\theta}$ implemented by this GCN can be expressed as:
% \begin{equation}
% f_\theta(\mathbf{A}, \mathbf{X})=\operatorname{softmax}\left(\hat{\mathbf{A}} \sigma\left(\hat{\mathbf{A}} \mathbf{X} \mathbf{W}^{(1)}\right) \mathbf{W}^{(2)}\right),
% \end{equation}
% where $\hat{\mathbf{A}}=\tilde{\mathbf{D}}^{-1 / 2}(\mathbf{A}+\mathbf{I}) \tilde{\mathbf{D}}^{-1 / 2}$ and $\tilde{\mathbf{D}}$ is the diagonal matrix of $\mathbf{A}+\mathbf{I}$ with $\tilde{\mathbf{D}}_{i i}=1+\sum_j \mathbf{A}_{i j}. \sigma$ is the activation function such as ReLU.

% \subsection{Graph Adversarial Attacks}
\noindent\textbf{Graph Adversarial Attacks.}
In the context of $G = (\mathbf{A},\mathbf{X})$ and a subset $V_m \subseteq V$ comprising victim nodes, where $y_i$ signifies the label for node $i$, the attacker's objective is to discern a perturbed graph denoted as $\tilde{G} = (\tilde{\mathbf{A}},\tilde{\mathbf{X}})$. The primary goal is to minimize the attack objective $\mathcal{L}_{attack}$. 
\begin{equation}
\min  \mathcal{L}_{\text {attack }}\left(f_\theta(\tilde{G})\right)=\sum\nolimits_{i \in V_m} \ell_{\text {attack }}\left(f_{\theta^*}(\tilde{G})_i, y_i\right) 
\text { s.t., }  \theta^*=\arg \min\nolimits_\theta \mathcal{L}_{\text {train }}\left(f_\theta\left(\hat{G}\right)\right),
\end{equation}
where $f_\theta$ indicates the model function of GNN,
$\mathcal{L}_{attack}$ represents the loss function for attacks, one option is to set $\mathcal{L}_{attack}$ = $-\mathcal{L}$, and $\hat{G}$ can be either $G$ or $\tilde{G}$. Here, $\tilde{G}$ is chosen from a constrained domain $\Psi(G)$. Given a fixed perturbation budget $\mathbb{D}$, a typical constraint for $\Psi(G)$ can be expressed as
$\|\tilde{\mathbf{A}}-\mathbf{A}\|_0+\|\tilde{\mathbf{X}}-\mathbf{X}\|_0 \leq \mathbb{D}$.
% \begin{equation}
% \|\tilde{\mathbf{A}}-\mathbf{A}\|_0+\|\tilde{\mathbf{X}}-\mathbf{X}\|_0 \leq \mathbb{D}.
% \end{equation}
This constraint implies that the perturbations introduced in the adjacency matrix $\tilde{\mathbf{A}}$ and node feature matrix $\tilde{\mathbf{X}}$ should be limited, and their combined $L_0$ should not exceed the specified budget $\mathbb{D}$. 

While graph adversarial attacks can perturb either node features or graph structures, the complexity of structural information has led the majority of existing adversarial attacks on graph data to focus on modifying graph structure, particularly through actions such as adding, deleting, or rewiring edges~\cite{jin2020graph,madry2017towards,geisler2021robustness,xu2019topology,chang2020restricted,ma2019attacking,entezari2020all,chen2021time,zhang2021projective}. 
For example, the \textbf{PGD} attack~\cite{madry2017towards} uses edge perturbation to overcome the challenge of attacking discrete graph structures via first-order optimization. In contrast, the \textbf{PRBCD} attack \cite{geisler2021robustness} addresses the high cost of adversarial attacks on large graphs with a sparsity-aware optimization approach.

On one hand, we explore the robustness against structural attacks. On the other hand, instead of targeting continuous features as in existing feature attack works, we adopt a direct approach by employing textual attacks to evaluate robustness, which remains a relatively unexplored direction.

\noindent\textbf{Textual Attack.}
%% taxonomy: character level, word level, sentence level
%% give definition of word level attack
For tasks on TAGs, the raw inputs are in text format and it can be hard for attackers to manipulate the encoded features directly, which makes the traditional feature attacking on graphs less practical. Therefore, the textual attack is used in this study to evaluate the robustness of LLMs enhanced graph features. Textual attacks can be performed on different levels like character level or sentence level according to the target to be perturbed. In this work, we focus on word-level attacks, which can be defined as follows.

\textit{Word-level attacks.} Given a classifier $f$ that predicts labels $y \in Y$, the input $X$ is defined in a categorical space and each input is a sequence of $n$ words ${x_1, x_2, \cdots, x_n}$. Each word $x_i$ has a limited amount of substitution candidates, denoted as $S(x)$. To keep the perturbation as unnoticeable as possible, a constraint, usually $L_1$ or $L_2$ distance is applied. Finally, to fool classifier $f$ to the largest extent, the following objective is maximized: 
% $ \arg \max_{x' \in S(x)} L(f(x'), y)  \text { s.t.,} \quad ||x' - x||_2 < \epsilon $,
\begin{equation}
\label{eq4}
\begin{array}{ll}
    \arg \max_{x' \in S(x)} L(f(x'), y)  \\
    \text { s.t.,} \quad ||x' - x||_2 < \epsilon, \\
\end{array}
\end{equation}
where $L$ is the loss function between original prediction $y$ and prediction $f(x')$ after perturbation. $\epsilon$ is the budget of the perturbation, which keeps the original and the perturbed sample as close as possible. As the loss function above is maximized, we will get a perturbed sample $x'$ that makes the classifier $f$ generate a prediction far from the original output.
For example, \textbf{SemAttack} \cite{wang2022semattack} generates natural adversarial text by employing various semantic perturbation functions.

\noindent\textbf{Large Language Models on Graphs.}
In recent years, remarkable progress has been achieved in the field of Large Language Models (LLMs), with notable contributions from transformative architectures such as Transformers~\cite{reimers2019sentence}, BERT~\cite{devlin2018bert}, Sentence-BERT (SBert)~\cite{reimers2019sentence} E5~\cite{wang2022text}, GPT~\cite{radford2018improving}, LLaMA~\cite{touvron2023llama} and their variants. These LLMs can be applied to graph-related tasks. The collaboration between Large Language Models and Graph Neural Networks (GNNs) can be mutually advantageous, leading to improved graph learning. The two most popular paradigms for applying LLMs to graphs are LLMs-as-Enhancers and LLMs-as-Predictors.
LLM-as-Enhancers aim to enhance the quality of node features through the assistance of LLMs.
% which can be categorized into two branches: explanation-based and embedding-based. 
TAPE~\cite{he2023harnessing} is a groundbreaking example of explanation-based enhancement, encouraging LLms to produce explanations and pseudo-labels for the augmentation of textual attributes. KEA~\cite{chen2024exploring} instructs LLMs to produce a compilation of knowledge entities, complete with text descriptions, and encodes them using fine-tuned pre-trained language models (PLMs). GLEM~\cite{zhao2022learning} considers pseudo labels generated by both PLMs and GNNs and incorporates them into a variational EM framework.
The fundamental concept of the LLMs-as-Predictors are to leverage LLMs for making predictions in graph-related tasks. InstructGLM~\cite{ye2023natural} formulates a set of scalable prompts grounded in the maximum hop level and fine-tunes LLMs to output predicted labels directly. GraphLLM~\cite{chai2023graphllm} derives the graph-enhanced prefix
from the graph representation. This method boosts the LLM’s capability in conducting graph reasoning tasks by graph-enhanced prefix tuning. The GraphGPT~\cite{tang2023graphgpt} framework aligns LLMs with graph structural knowledge using a paradigm of graph instruction tuning.

% \section{Threat Model}
% \label{sec:sup-threat}
% We describe the characteristics of the graph adversarial attacks we developed, including both structural and textual attacks, from the following aspects. \textit{(1) Adversary's Goal}: The primary objective is focused on evasion attacks, where the adversary seeks to manipulate the input graph data at inference time to cause the model to make incorrect predictions. In this scenario, the adversaries do not have the authority to change the classifier or its parameters.
% \textit{(2) Adversary’s Knowledge:} \zw{The attacks are designed under white-box and grey-box frameworks, meaning that the adversary either possesses complete knowledge of the model architecture and parameters or not respectively}. \textit{(3) Victim Models:} 
% The targets of these attacks include LLMs-as-Predictors and LLMs-as-Enhancers, of which the details will be thoroughly elaborated in the following subsections. 

\section{Benchmark Datasets}
\label{sec:sup-dataset}
To comprehensively and effectively assess the robustness of LLMs in graph learning, we present six text-attributed graphs that offer original textual sentences. For instance, renowned citation graphs such as Cora~\cite{sen2008collective}, PubMed~\cite{sen2008collective}, Citeseer~\cite{sen2008collective}, and ogbn-arxiv (Arxiv)~\cite{hu2020open} fall under the category of TAGs. These datasets extract node attributes from textual information, including titles and abstracts of papers. WikiCS~\cite{mernyei2020wiki} serves as a Wikipedia-based dataset for benchmarking Graph Neural Networks. It comprises 10 classes corresponding to branches of computer science, demonstrating high connectivity. Node features are derived from the text of the respective articles. Moreover, we employ the History~\cite{yan2023comprehensive} dataset sourced from Amazon, where node attributes originate from book titles and descriptions. For instance, "Description: Collection of Poetry; Title: The golden treasury of poetry". All datasets are utilized with a low labeling rate split, following the setting described in KEA~\cite{chen2024exploring}. The statistics of datasets are reported in Table~\ref{tab:statistic}.
The licenses of these datasets are MIT License.
\begin{table}[h]
\centering

 \caption{Statistics of datasets. 
 }
 \label{tab:statistic}
 \resizebox{1\textwidth}{!}{%
\begin{tabular}{ccccccc}
\toprule
 & Cora & Citeseer & Pubmed & WikiCS & History & ogbn-arxiv  \\
 \midrule
\#Nodes & \multicolumn{1}{r}{2,708} & \multicolumn{1}{r}{3,327} & \multicolumn{1}{r}{19,717} & \multicolumn{1}{r}{11,701} & \multicolumn{1}{r}{41,551}  & \multicolumn{1}{r}{169,343}  \\
\#Edges & \multicolumn{1}{r}{5,429} & \multicolumn{1}{r}{4,732} & \multicolumn{1}{r}{44,338} & \multicolumn{1}{r}{216,123} & \multicolumn{1}{r}{358,574} & \multicolumn{1}{r}{1,166,243}  \\
\#Class & \multicolumn{1}{r}{7} & \multicolumn{1}{r}{6} & \multicolumn{1}{r}{3} & \multicolumn{1}{r}{10} & \multicolumn{1}{r}{12} & \multicolumn{1}{r}{40} \\
\#Domain& \multicolumn{1}{r}{Academic}&\multicolumn{1}{r}{Academic} &\multicolumn{1}{r}{Academic} &\multicolumn{1}{r}{Wikipedia} & \multicolumn{1}{r}{E-commerce} & \multicolumn{1}{r}{Academic}\\
\bottomrule
\end{tabular}}
\label{table:app_data}
\end{table}

\section{Experimental Settings}
\label{sec:sup-setting}
All algorithms in our benchmark are implemented by PyTorch~\cite{paszke2019pytorch}. All experiments are conducted on a Linux server with GPU (NVIDIA RTX A6000
48Gb and Tesla V100 32Gb), using PyTorch 2.0.0, PyTorch Geometric 2.4.0~\cite{fey2019fast} and Python 3.10.13.
We train each model using binary cross-entropy loss, optimized with the Adam optimizer. We use PGD and PRBCD to attack the structures on graphs, which are implemented by DeepRobust~\cite{li2020deeprobust}. For hyperparameter,  a hidden size of 256 is used uniformly across all datasets. For the sake of reproducibility,
the seeds of random numbers are set to the same. For all the datasets and models, we tune the following hyper-parameters:  learning rate: $lr \in\{0.01, 0.001\}$, weight decay: $\lambda \in\{1e-4, 5e-4\}$.
% \begin{table}[h!]
% \centering
% \caption{Dataset Licenses}
% \begin{tabular}{@{}c|c@{}}
% \toprule
% \textbf{Datasets} & \textbf{License} \\ \midrule
% Cora & MIT License \\
% Citeseer & MIT License \\
% Pubmed & MIT License \\
% ogbn-arxiv & MIT License \\
% History & MIT License \\
% wikics & MIT License \\
% \bottomrule
% \end{tabular}
% \end{table}

\section{Results of LLMs-as-Enhancers against attacks on graph structures}
\label{sec:sup-enhancer4structure}
To answer \textbf{Q1}, we conduct experiments to examine the robustness against structural attack for LLMs-as-Enhancers. The results are reported in Table~\ref{tab:enhancer-structure2} and Table~\ref{tab:enhancer-structure1}, which include additional details on the standard deviation (std) compared to Table~\ref{tab:enhancer-structure5}.
\begin{table}[htbp]
\caption{The performance of LLMs-as-Enhancers against \textbf{5\%} attacks on graph structures}
\label{tab:enhancer-structure2}
\resizebox{1\textwidth}{!}{%
\begin{tabular}{@{}c|ccc|ccc|ccc@{}}
\toprule
         & \multicolumn{3}{c|}{Cora}                                                                 & \multicolumn{3}{c|}{Pudmed}                                                             & \multicolumn{3}{c}{Arxiv}                                                                \\ \midrule
Feature       & \multicolumn{1}{c|}{Clean}           & \multicolumn{1}{c|}{5\%  Attack}      & Gap          & \multicolumn{1}{c|}{Clean}           & \multicolumn{1}{c|}{5\%  Attack}      & Gap           & \multicolumn{1}{c|}{Clean}           & \multicolumn{1}{c|}{5\%  Attack}      & Gap           \\ \midrule
BOW      & \multicolumn{1}{c|}{78.49 ± 1.13} & \multicolumn{1}{c|}{73.71 ± 1.10} & 03.06\% &  \multicolumn{1}{c|}{74.69 ± 2.07} & \multicolumn{1}{c|}{72.40 ± 1.89} & 15.88\% & \multicolumn{1}{c|}{50.99 ± 2.15} & \multicolumn{1}{c|}{42.35 ± 2.64} & 16.94\% \\
TFIDF    & \multicolumn{1}{c|}{81.46 ± 1.21} & \multicolumn{1}{c|}{77.23 ± 1.07} & 05.19\% & \multicolumn{1}{c|}{76.86 ± 1.34} & \multicolumn{1}{c|}{74.68 ± 1.40} & 02.84\% & \multicolumn{1}{c|}{48.39 ± 1.15} & \multicolumn{1}{c|}{43.32 ± 1.27} &10.48\% \\
SBert    & \multicolumn{1}{c|}{81.99 ± 0.76} & \multicolumn{1}{c|}{79.18 ± 0.71} & 03.43\%  & \multicolumn{1}{c|}{78.71 ± 1.17} & \multicolumn{1}{c|}{74.68 ± 1.40} & 05.12\% & \multicolumn{1}{c|}{52.51 ± 0.87} & \multicolumn{1}{c|}{47.59 ± 1.72} & 09.37\%\\
E5       & \multicolumn{1}{c|}{83.17 ± 0.73} & \multicolumn{1}{c|}{80.43 ± 0.59} & 03.29\%  & \multicolumn{1}{c|}{81.83 ± 1.16} & \multicolumn{1}{c|}{79.39 ± 1.06} & 02.98\% & \multicolumn{1}{c|}{57.04 ± 1.77} & \multicolumn{1}{c|}{48.15 ± 1.71} & 15.59\% \\
Llama    & \multicolumn{1}{c|}{78.13 ± 1.07} & \multicolumn{1}{c|}{70.91 ± 1.02} & 09.24\%  & \multicolumn{1}{c|}{77.65 ± 0.74} & \multicolumn{1}{c|}{76.68 ± 0.94} & 01.25\% & \multicolumn{1}{c|}{58.04 ± 1.79} & \multicolumn{1}{c|}{51.76 ± 1.95} & 10.82\%\\
Angle-Llama & \multicolumn{1}{c|}{80.28 ± 1.42} & \multicolumn{1}{c|}{80.23 ± 0.92} & 0.06\%  & \multicolumn{1}{c|}{75.15 ± 2.50} & \multicolumn{1}{c|}{74.65 ± 1.45} & 0.67\% & \multicolumn{1}{c|}{58.53 ± 1.56} & \multicolumn{1}{c|}{51.46 ± 3.33} & 12.08\%\\
Explanation       & \multicolumn{1}{c|}{82.79 ± 1.17} & \multicolumn{1}{c|}{80.47 ± 1.28} & 02.80\% & \multicolumn{1}{c|}{88.84 ± 0.34} & \multicolumn{1}{c|}{87.32 ± 0.38} & 01.52\% & \multicolumn{1}{c|}{54.37 ± 5.51}             & \multicolumn{1}{c|}{47.19 ± 1.13}             &         13.21\%          \\
Ensemble       & \multicolumn{1}{c|}{82.57 ± 1.48} & \multicolumn{1}{c|}{80.67 ± 1.39} & 02.30\%  & \multicolumn{1}{c|}{83.59 ± 0.81} & \multicolumn{1}{c|}{82.55 ± 1.41} & 01.24\% & \multicolumn{1}{c|}{59.90 ± 2.93}             & \multicolumn{1}{c|}{54.33 ± 2.43}             &  09.30\%                \\
Llama-FT & \multicolumn{1}{c|}{78.54 ± 1.53} & \multicolumn{1}{c|}{78.51 ± 0.88} & 0\%  & \multicolumn{1}{c|}{77.40 ± 0.78} & \multicolumn{1}{c|}{76.55 ± 0.85} &             1.10\%      & \multicolumn{1}{c|}{52.46 ± 0.84 } & \multicolumn{1}{c|}{47.59 ± 1.71} & 09.28\%\\ \bottomrule
\toprule
         & \multicolumn{3}{c|}{Wikics}                                                                 & \multicolumn{3}{c|}{History}                                                             & \multicolumn{3}{c}{Citeseer}                                                                \\ \midrule
Feature       & \multicolumn{1}{c|}{Clean}           & \multicolumn{1}{c|}{5\%  Attack}      & Gap          & \multicolumn{1}{c|}{Clean}           & \multicolumn{1}{c|}{5\%  Attack}      & Gap          & \multicolumn{1}{c|}{Clean}           & \multicolumn{1}{c|}{5\%  Attack}      & Gap          \\ \midrule
BOW      & \multicolumn{1}{c|}{74.92 ± 0.05} & \multicolumn{1}{c|}{61.46 ± 0.32} & 17.95\% & \multicolumn{1}{c|}{51.25 ± 4.80} & \multicolumn{1}{c|}{49.04 ± 5.80} & 04.31\%  &\multicolumn{1}{c|}{70.74 ± 0.72} & \multicolumn{1}{c|}{68.84 ± 0.82} & 02.69\% \\
TFIDF & \multicolumn{1}{c|}{75.96 ± 0.14} & \multicolumn{1}{c|}{62.03 ± 0.31} &18.34\% &\multicolumn{1}{c|} {58.38 ± 3.25} & \multicolumn{1}{c|}{57.57 ± 1.69} &01.39\%  & \multicolumn{1}{c|}{73.02 ± 0.66} & \multicolumn{1}{c|}{71.40 ± 0.78} & 02.22\% \\ 
SBert & \multicolumn{1}{c|}{75.89 ± 1.77} & \multicolumn{1}{c|}{64.44 ± 2.33} & 15.09\% & \multicolumn{1}{c|}{65.59 ± 1.60} & \multicolumn{1}{c|}{65.60 ± 1.48} & 0\%  & \multicolumn{1}{c|}{74.94 ± 0.88} & \multicolumn{1}{c|}{72.60 ± 0.81} & 03.14\%\\
E5 & \multicolumn{1}{c|}{76.71 ± 1.74} & \multicolumn{1}{c|}{63.01 ± 2.13} & 17.86\% & \multicolumn{1}{c|}{66.96 ± 2.30} & \multicolumn{1}{c|}{64.69 ± 2.10} & 03.39\% & \multicolumn{1}{c|}{75.14 ± 0.93} & \multicolumn{1}{c|}{72.29 ± 1.27} & 03.79\%\\
Llama & \multicolumn{1}{c|}{79.72 ± 0.35} & \multicolumn{1}{c|}{70.91 ± 1.02} & 11.05\% & \multicolumn{1}{c|}{66.73 ± 3.53} & \multicolumn{1}{c|}{61.64 ± 3.21} & 07.63\% & \multicolumn{1}{c|}{67.48 ± 3.40} & \multicolumn{1}{c|}{70.82 ± 1.07} & 04.95\% \\
Angle-Llama & \multicolumn{1}{c|}{73.33 ± 1.77} & \multicolumn{1}{c|}{67.00 ± 3.62} &08.63\% & \multicolumn{1}{c|}{66.52 ± 1.69} & \multicolumn{1}{c|}{63.47 ± 3.14} & 04.59\% & \multicolumn{1}{c|}{71.71 ± 1.73} & \multicolumn{1}{c|}{68.46 ± 4.92} & 04.53\% \\ 
Llama-FT & \multicolumn{1}{c|}{79.69 ± 0.39} & \multicolumn{1}{c|}{70.11 ± 1.03} & 12.02\% & \multicolumn{1}{c|}{64.53 ± 2.02} & \multicolumn{1}{c|}{62.99 ± 1.44}             & 02.39\% & \multicolumn{1}{c|}{69.70 ± 2.18} & \multicolumn{1}{c|}{68.62 ± 3.15} & 01.55\%                   \\
\bottomrule
\end{tabular}}
\end{table}

\begin{table}[htbp]
\caption{The performance of LLMs-as-Enhancers against \textbf{25\%} attacks on graph structures}
\label{tab:enhancer-structure1}
\resizebox{1\textwidth}{!}{%
\begin{tabular}{@{}c|ccc|ccc|ccc@{}}
\toprule
         & \multicolumn{3}{c|}{Cora}                                                                 & \multicolumn{3}{c|}{Pudmed}                                                             & \multicolumn{3}{c}{Arxiv}                                                                \\ \midrule
Feature       & \multicolumn{1}{c|}{Clean}           & \multicolumn{1}{c|}{25\%  Attack}      & Gap          & \multicolumn{1}{c|}{Clean}           & \multicolumn{1}{c|}{25\%  Attack}      & Gap          & \multicolumn{1}{c|}{Clean}           & \multicolumn{1}{c|}{25\%  Attack}      & Gap          \\ \midrule
BOW      & \multicolumn{1}{c|}{78.49 ± 1.13} & \multicolumn{1}{c|}{60.72 ± 1.92} & 22.64\% &  \multicolumn{1}{c|}{74.69 ± 2.07} & \multicolumn{1}{c|}{62.83 ± 1.61} & 15.88\% & \multicolumn{1}{c|}{50.99 ± 2.15} & \multicolumn{1}{c|}{18.95 ± 1.63} & 62.85\% \\
TFIDF    & \multicolumn{1}{c|}{81.46 ± 1.21} & \multicolumn{1}{c|}{67.53 ± 0.82} & 17.10\%& \multicolumn{1}{c|}{76.86 ± 1.34} & \multicolumn{1}{c|}{66.18 ± 1.44} & 13.90\% & \multicolumn{1}{c|}{48.39 ± 1.15} & \multicolumn{1}{c|}{24.36 ± 2.10} &49.67\% \\
SBert    & \multicolumn{1}{c|}{81.99 ± 0.76} & \multicolumn{1}{c|}{72.14 ± 2.13} & 12.01\%  & \multicolumn{1}{c|}{78.71 ± 1.17} & \multicolumn{1}{c|}{69.24 ± 1.85} & 12.03\% & \multicolumn{1}{c|}{52.51 ± 0.87} & \multicolumn{1}{c|}{29.24 ± 3.27} & 44.32\%\\
E5       & \multicolumn{1}{c|}{83.17 ± 0.73} & \multicolumn{1}{c|}{69.84 ± 0.55} & 16.03\%  & \multicolumn{1}{c|}{81.83 ± 1.16} & \multicolumn{1}{c|}{71.06 ± 1.12} & 13.15\% & \multicolumn{1}{c|}{57.04 ± 1.77} & \multicolumn{1}{c|}{25.24 ± 2.77} & 55.75\% \\
Llama    & \multicolumn{1}{c|}{78.13 ± 1.07} & \multicolumn{1}{c|}{69.99 ± 1.92} & 10.42\%  & \multicolumn{1}{c|}{77.65 ± 0.72} & \multicolumn{1}{c|}{74.62 ± 1.64} & 03.89\% & \multicolumn{1}{c|}{58.04 ± 1.79} & \multicolumn{1}{c|}{31.67 ± 3.09} & 45.43\%\\
Angle-Llama & \multicolumn{1}{c|}{80.28 ± 1.42} & \multicolumn{1}{c|}{76.75 ± 1.33} & 04.40\%  & \multicolumn{1}{c|}{75.15 ± 2.50} & \multicolumn{1}{c|}{72.72 ± 3.00} & 03.23\% & \multicolumn{1}{c|}{58.53 ± 1.56} & \multicolumn{1}{c|}{31.73 ± 4.80} & 45.79\%\\
Explanation       & \multicolumn{1}{c|}{82.79 ± 1.17} & \multicolumn{1}{c|}{71.32 ± 2.28} & 13.85\% & \multicolumn{1}{c|}{88.84 ± 0.34} & \multicolumn{1}{c|}{82.11 ± 0.72} & 07.57\% & \multicolumn{1}{c|}{54.37 ± 5.51}             & \multicolumn{1}{c|}{21.61 ± 3.10}             &         60.25\%          \\
Ensemble       & \multicolumn{1}{c|}{82.57 ± 1.48} & \multicolumn{1}{c|}{74.89 ± 0.93} & 09.30\%  & \multicolumn{1}{c|}{83.59 ± 0.81} & \multicolumn{1}{c|}{77.85 ± 0.50} & 06.87\% & \multicolumn{1}{c|}{59.90 ± 2.93}             & \multicolumn{1}{c|}{32.82 ± 3.19}             &  45.21\%                \\
Llama-FT & \multicolumn{1}{c|}{78.54 ± 1.53} & \multicolumn{1}{c|}{70.72 ± 4.15} & 09.96\%  & \multicolumn{1}{c|}{77.40 ± 0.78} & \multicolumn{1}{c|}{75.63 ± 2.15} &             2.29\%      & \multicolumn{1}{c|}{52.46 ± 0.84 } & \multicolumn{1}{c|}{29.19 ± 3.19} & 44.36\%\\ \bottomrule
\toprule
         & \multicolumn{3}{c|}{Wikics}                                                                 & \multicolumn{3}{c|}{History}                                                             & \multicolumn{3}{c}{Citeseer}                                                                \\ \midrule
Feature       & \multicolumn{1}{c|}{Clean}           & \multicolumn{1}{c|}{25\%  Attack}      & Gap          & \multicolumn{1}{c|}{Clean}           & \multicolumn{1}{c|}{25\%  Attack}      & Gap          & \multicolumn{1}{c|}{Clean}           & \multicolumn{1}{c|}{25\%  Attack}      & Gap          \\ \midrule
BOW      & \multicolumn{1}{c|}{74.92 ± 0.05} & \multicolumn{1}{c|}{45.34 ± 0.31} & 39.45\% & \multicolumn{1}{c|}{51.25 ± 4.80} & \multicolumn{1}{c|}{38.93 ± 5.53} & 24.02\%  &\multicolumn{1}{c|}{70.74 ± 0.72} & \multicolumn{1}{c|}{61.76 ± 1.44} & 12.69\% \\
TFIDF & \multicolumn{1}{c|}{75.96 ± 0.14} & \multicolumn{1}{c|}{45.74 ± 0.51} &39.76\% &\multicolumn{1}{c|} {58.38 ± 3.25} & \multicolumn{1}{c|}{22.66 ± 1.69} &61.15\%  & \multicolumn{1}{c|}{73.02 ± 0.66} & \multicolumn{1}{c|}{64.36 ± 0.72} & 11.86\% \\ 
SBert & \multicolumn{1}{c|}{75.89 ± 1.77} & \multicolumn{1}{c|}{50.19 ± 3.20} & 33.86\% & \multicolumn{1}{c|}{65.59 ± 1.60} & \multicolumn{1}{c|}{56.34 ± 2.41} & 14.10\%  & \multicolumn{1}{c|}{74.94 ± 0.88} & \multicolumn{1}{c|}{64.51 ± 1.50} & 13.92\%\\
E5 & \multicolumn{1}{c|}{76.71 ± 1.74} & \multicolumn{1}{c|}{46.38 ± 4.50} & 39.55\% & \multicolumn{1}{c|}{66.96 ± 2.30} & \multicolumn{1}{c|}{55.37 ± 1.80} & 17.30\% & \multicolumn{1}{c|}{75.14 ± 0.93} & \multicolumn{1}{c|}{64.38 ± 1.83} & 14.31\%\\
Llama & \multicolumn{1}{c|}{79.72 ± 0.35} & \multicolumn{1}{c|}{59.23 ± 1.47} & 25.71\% & \multicolumn{1}{c|}{66.73 ± 3.53} & \multicolumn{1}{c|}{54.54 ± 1.74} & 18.26\% & \multicolumn{1}{c|}{67.48 ± 3.40} & \multicolumn{1}{c|}{64.69 ± 2.99} & 04.13\% \\
Angle-Llama & \multicolumn{1}{c|}{73.33 ± 1.77} & \multicolumn{1}{c|}{56.75 ± 4.65} &22.61\% & \multicolumn{1}{c|}{66.52 ± 1.69} & \multicolumn{1}{c|}{52.75 ± 3.66} & 20.71\% & \multicolumn{1}{c|}{71.71 ± 1.73} & \multicolumn{1}{c|}{68.61 ± 3.61} & 04.32\% \\ 
Llama-FT & \multicolumn{1}{c|}{79.69 ± 0.39} & \multicolumn{1}{c|}{60.13 ± 0.31} & 24.56\% & \multicolumn{1}{c|}{64.53 ± 2.02} & \multicolumn{1}{c|}{53.34 ± 2.47}             & 17.36\% & \multicolumn{1}{c|}{69.70 ± 2.18} & \multicolumn{1}{c|}{66.17 ± 3.62} & 05.07\%                   \\
\bottomrule
\end{tabular}}
\end{table}

\section{Results of LLMs-as-Enhancers against attacks on texts}
\label{sec:sup-enhancer4text}
To address \textbf{Q2}, we perform a white-box evasion attack on LLMs-as-Enhancers, specifically targeting textual attributes. The results are shown in Table~\ref{tab:sup-enhancer-textual_full}, which includes additional details on the standard deviation (std) compared to Table~\ref{tab:enhancer-textual}.
\begin{table}[h]
\centering

\caption{Attack success rate results of LLMs-as-Enhancers against \textbf{5\%} textual attack.}
\label{tab:sup-enhancer-textual_full}
\resizebox{1\textwidth}{!}{%
\begin{tabular}{@{}c|cc|cc|cc@{}}
\toprule
 & \multicolumn{2}{c|}{Cora} & \multicolumn{2}{c|}{PubMed} & \multicolumn{2}{c}{Arxiv} \\ \midrule
Features      & \multicolumn{1}{c|}{MLP}    & \multicolumn{1}{c|}{GCN}   & \multicolumn{1}{c|}{MLP}    & \multicolumn{1}{c|}{GCN}    & \multicolumn{1}{c|}{MLP}    & \multicolumn{1}{c}{GCN}  \\
\midrule
BOW     & \multicolumn{1}{c|}{62.11±4.48}  & \multicolumn{1}{c|}{9.27±2.97}   & \multicolumn{1}{c|}{41.00±7.86}  & \multicolumn{1}{c|}{8.58±2.45}  & \multicolumn{1}{c|}{72.69±6.93}   & \multicolumn{1}{c}{15.19±4.86}   \\ \midrule
Sbert   & \multicolumn{1}{c|}{73.18±2.02}  & \multicolumn{1}{c|}{14.76±2.48}  & \multicolumn{1}{c|}{45.36±3.88}  & \multicolumn{1}{c|}{9.32±3.89}  & \multicolumn{1}{c|}{82.69±10.88}  & \multicolumn{1}{c}{11.17±1.39}   \\
E5      & \multicolumn{1}{c|}{65.29±6.42}  & \multicolumn{1}{c|}{10.53±1.86}  & \multicolumn{1}{c|}{35.62±5.64}  & \multicolumn{1}{c|}{8.76±2.59}  & \multicolumn{1}{c|}{81.92±8.76}   & \multicolumn{1}{c}{15.22±4.18}    \\
LLaMA   & \multicolumn{1}{c|}{56.49±7.34}  & \multicolumn{1}{c|}{12.58±0.85}  & \multicolumn{1}{c|}{19.66±4.89}  & \multicolumn{1}{c|}{4.95±2.59}  & \multicolumn{1}{c|}{67.87±13.10}  & \multicolumn{1}{c}{6.05±2.56}    \\
LLaMA-FT& \multicolumn{1}{c|}{40.10±12.16} & \multicolumn{1}{c|}{4.37±1.94}   & \multicolumn{1}{c|}{16.69±5.57}  & \multicolumn{1}{c|}{3.27±1.65}  & \multicolumn{1}{c|}{67.71±9.83}   & \multicolumn{1}{c}{6.08±2.00}  \\ 
\bottomrule
\toprule

 & \multicolumn{2}{c|}{Wikics} & \multicolumn{2}{c|}{History} & \multicolumn{2}{c}{Citeseer} \\ \midrule
Features      & \multicolumn{1}{c|}{MLP}    & \multicolumn{1}{c|}{GCN}   & \multicolumn{1}{c|}{MLP}    & \multicolumn{1}{c|}{GCN}    & \multicolumn{1}{c|}{MLP}    & \multicolumn{1}{c}{GCN}  \\
\midrule
BOW       & \multicolumn{1}{c|}{67.85±4.88}  & \multicolumn{1}{c|}{3.82±1.46}   & \multicolumn{1}{c|}{74.53±10.13}  & \multicolumn{1}{c|}{18.40±6.00}  & \multicolumn{1}{c|}{68.80±6.99}  & \multicolumn{1}{c}{18.98±4.67}   \\ \midrule
Sbert     & \multicolumn{1}{c|}{62.12±4.32}  & \multicolumn{1}{c|}{9.08±2.34}   & \multicolumn{1}{c|}{76.73±8.79}   & \multicolumn{1}{c|}{13.41±3.86}  & \multicolumn{1}{c|}{66.33±6.09}  & \multicolumn{1}{c}{16.93±3.31}   \\
E5        & \multicolumn{1}{c|}{65.99±5.05}  & \multicolumn{1}{c|}{6.64±2.47}   & \multicolumn{1}{c|}{61.51±15.43}  & \multicolumn{1}{c|}{6.54±1.04}   & \multicolumn{1}{c|}{57.10±6.25}  & \multicolumn{1}{c}{14.69±3.73 }  \\
LLaMA     & \multicolumn{1}{c|}{22.17±4.96}  & \multicolumn{1}{c|}{4.17±0.80}   & \multicolumn{1}{c|}{65.07±9.50}   & \multicolumn{1}{c|}{15.92±8.47}  & \multicolumn{1}{c|}{46.81±14.07} & \multicolumn{1}{c}{13.77±5.34 }  \\
LLaMA-FT  & \multicolumn{1}{c|}{30.00±2.07}  & \multicolumn{1}{c|}{2.98±1.83}   & \multicolumn{1}{c|}{56.98±14.55}  & \multicolumn{1}{c|}{6.91±3.64}   & \multicolumn{1}{c|}{38.46±11.19} & \multicolumn{1}{c}{6.58±2.97}  \\
\bottomrule
\end{tabular}}
\end{table}

\section{Results of LLMs-as-Predictors against attacks on graph structures}
\label{sec:sup-predictor4structure}
To address \textbf{Q3}, we explore the robustness of predictors against graph structure attacks, using GPT-3.5 as the predictor. The results are presented in Figure~\ref{fig:predictor-structual}.
We evaluate the model under various few-shot settings and the zero-shot setting with the summary of two-hop neighbors. The experiment shows that GPT-3.5 demonstrates the strongest robustness in the zero-shot setting, with performance barely dropping even under significant perturbations. This resilience is likely due to the neighbor sampling and summarizing processes, which mitigate the noise introduced by structural changes. Although GAT also uses the attacked graph structure migrated from GCN, its accuracies decline much faster compared to GPT-3.5.
\begin{figure}[h]
  \centering
  \includegraphics[width=1\columnwidth]{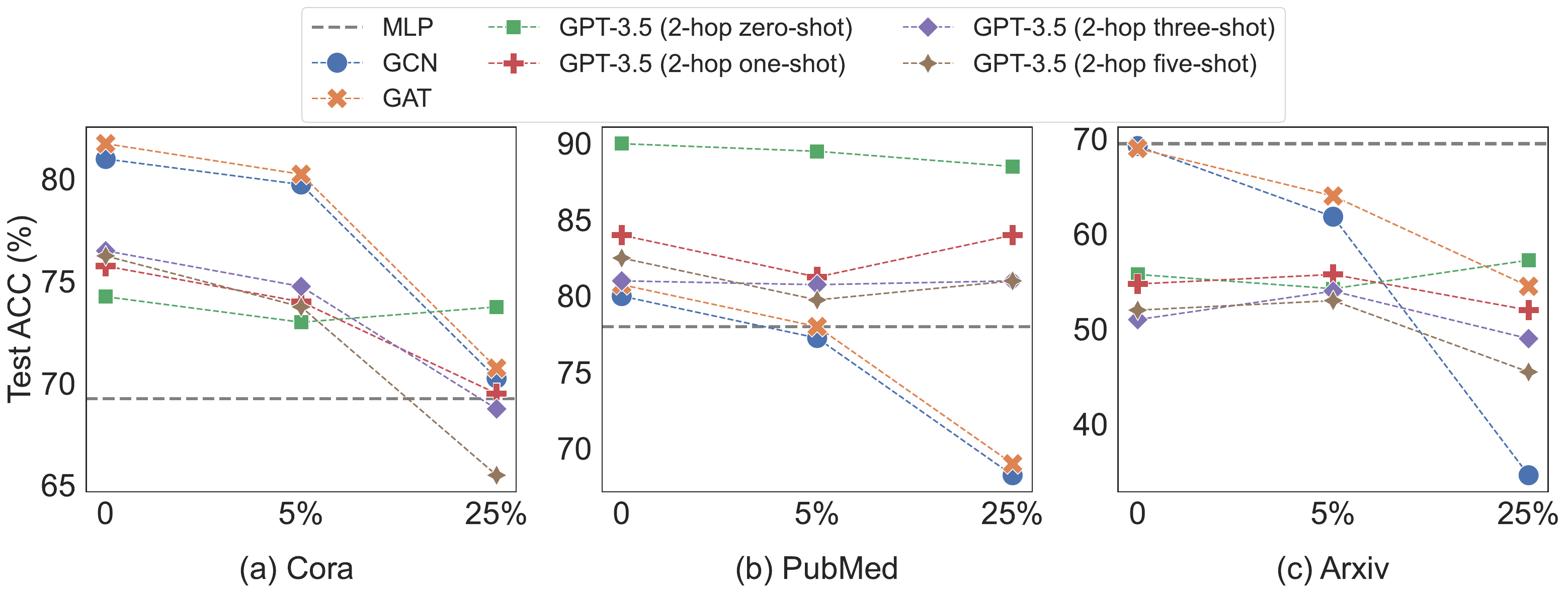} 
  \caption{The performance of LLMs-as-Predictors against structural attacks, evaluated by accuracy.}
  \label{fig:predictor-structual}
  \vskip -1.2em
\end{figure}

\section{Results of LLMs-as-Predictors against attacks on texts}
\label{sec:sup-predictor4text}
To address \textbf{Q4}, we explore the performance of GPT-3.5 with perturbed texts as inputs. The results are presented in Table~\ref{tab:sup-predictor-textual}, showing that GPT-3.5 demonstrates strong robustness compared to MLP but does not surpass GCN and GAT in all few-shot settings. On the other hand, GAT exhibits the strongest robustness, followed by GCN. However, on the Cora and Pubmed datasets, the Attack Success Rate of GPT-3.5 is comparable to GCN, ranging from 0.55\% to 1.70\%.
\begin{table*}[tp]
\centering
\caption{The performance of the predictor GPT-3.5 against \textbf{5\%} textual attacks. The bold font is used to highlight the lowest Attack Success Rate.}
\label{tab:sup-predictor-textual}
\resizebox{1\textwidth}{!}{%
\begin{tabular}{l|cc|cc|cc}
\toprule
Dataset   & \multicolumn{2}{c|}{Cora} & \multicolumn{2}{c|}{Pubmed} & \multicolumn{2}{c}{Arxiv} \\ \hline
Model  & \multicolumn{1}{c|}{Clean ACC}   & \multicolumn{1}{c|}{ASR}    & \multicolumn{1}{c|}{Clean ACC}   & \multicolumn{1}{c|}{ASR}   & \multicolumn{1}{c|}{Clean ACC}           & \multicolumn{1}{c}{ASR} \\ \midrule
MLP   & \multicolumn{1}{c|}{69.25}      & \multicolumn{1}{c|}{35.97}        & \multicolumn{1}{c|}{78.00}           & \multicolumn{1}{c|}{14.74}       & \multicolumn{1}{c|}{69.50}     &\multicolumn{1}{c}{30.10}    \\
GCN   & \multicolumn{1}{c|}{81.00}         & \multicolumn{1}{c|}{4.30}      & \multicolumn{1}{c|}{80.00}           & \multicolumn{1}{c|}{3.20}       & \multicolumn{1}{c|}{69.25}       & \multicolumn{1}{c}{5.42}      \\
GAT   & \multicolumn{1}{c|}{81.75}      & \multicolumn{1}{c|}{\textbf{2.99}}          & \multicolumn{1}{c|}{80.75}        & \multicolumn{1}{c|}{\textbf{2.23}}        & \multicolumn{1}{c|}{69.00}      & \multicolumn{1}{c}{\textbf{3.97}}   \\
GPT-3.5 (2-hop zero-shot)  & \multicolumn{1}{c|}{74.25}      & \multicolumn{1}{c|}{10.03}      & \multicolumn{1}{c|}{90.00}           & \multicolumn{1}{c|}{5.53}       & \multicolumn{1}{c|}{55.75}       & \multicolumn{1}{c}{13.33}     \\
GPT-3.5 (2-hop one-shot)   & \multicolumn{1}{c|}{75.75}      & \multicolumn{1}{c|}{5.59}     & \multicolumn{1}{c|}{84.00}           & \multicolumn{1}{c|}{3.75}       & \multicolumn{1}{c|}{54.75}       & \multicolumn{1}{c}{18.65 }    \\
GPT-3.5 (2-hop three-shot) & \multicolumn{1}{c|}{76.50}       & \multicolumn{1}{c|}{5.59}       & \multicolumn{1}{c|}{81.00}           & \multicolumn{1}{c|}{5.29}       & \multicolumn{1}{c|}{51.00}          & \multicolumn{1}{c}{19.10}          \\
GPT-3.5 (2-hop five-shot)  & \multicolumn{1}{c|}{76.25}      & \multicolumn{1}{c|}{6.72}       & \multicolumn{1}{c|}{82.50 }        & \multicolumn{1}{c|}{6.07}       & \multicolumn{1}{c|}{52.00}          & \multicolumn{1}{c}{21.24}     \\

\bottomrule
\end{tabular}}
\end{table*}

\section{Analysis for Textual Attack}
\label{sec:sup-analysis}
From the experiments on LLMs-as-Enhancers against textual attacks in Section \ref{sec:enhancer-text}, we observe that LLaMA performs significantly better than other smaller models, with fine-tuned LLaMA demonstrating the best robustness. Additionally, GCN shows much stronger robustness compared to MLP.

To explore the reasons behind these observations, we analyze the models from feature and structure perspectives. From the feature perspective, we use SBert and GCN as the victim model and use Latent Dirichlet Allocation to generate the distribution of three themes for each node. Then, we acquire the mean value for each theme. As shown in Fig.~\ref{fig:analysis_LDA}, the theme distributions for the successfully and failed attacked nodes are similar, indicating that there is not a strong relation between text content and robustness. From the structure perspective, we examine the relationship between the centrality of nodes and the attack success rate. As shown in Fig.~\ref{fig:analysis_centrality_pubmed}, we use SBert and GCN as the victim models and visualize the degree distributions of successfully attacked nodes versus failed attacked nodes.
% We observed from perspectives a, b, and c that 

It is evident that successfully attacked nodes often have smaller degrees, indicating that nodes with less structural information in the graph are more vulnerable to textual attacks. Furthermore, similar patterns are observed with eigenvector centrality and PageRank values.
\begin{figure}[h]
  \centering
  \includegraphics[width=1\columnwidth]{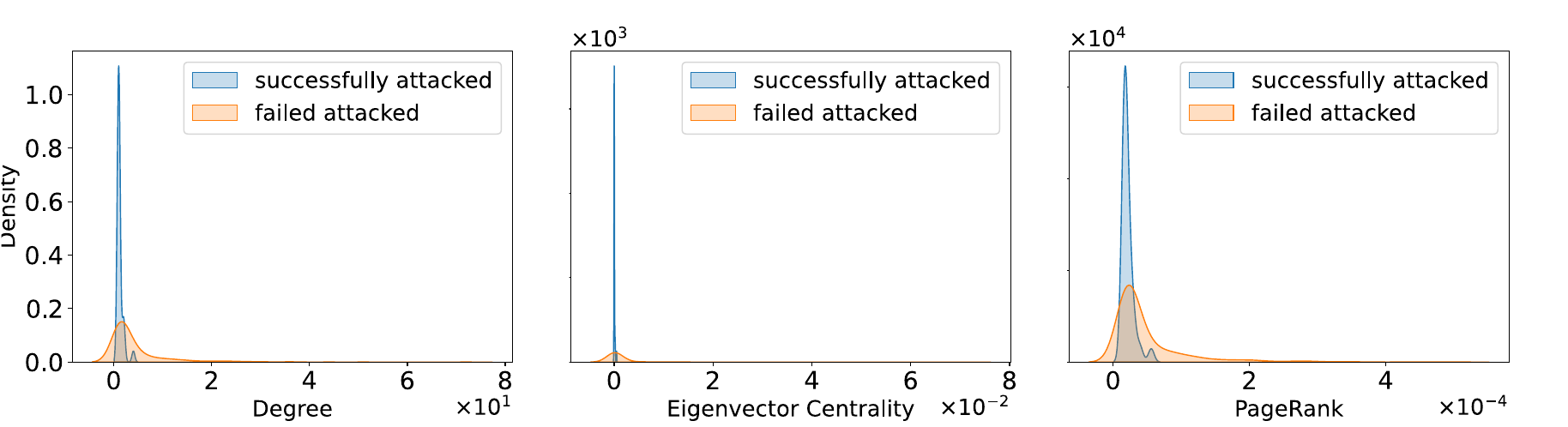} 
  \caption{Centrality distributions of the node being attacked successfully and unsuccessfully. We use Sentence-Bert as the victim model and gather all attacked results from the PubMed dataset.}
  \label{fig:analysis_centrality_pubmed}
\end{figure}

\begin{figure}[tp]
  \centering
\includegraphics[width=0.85\columnwidth]{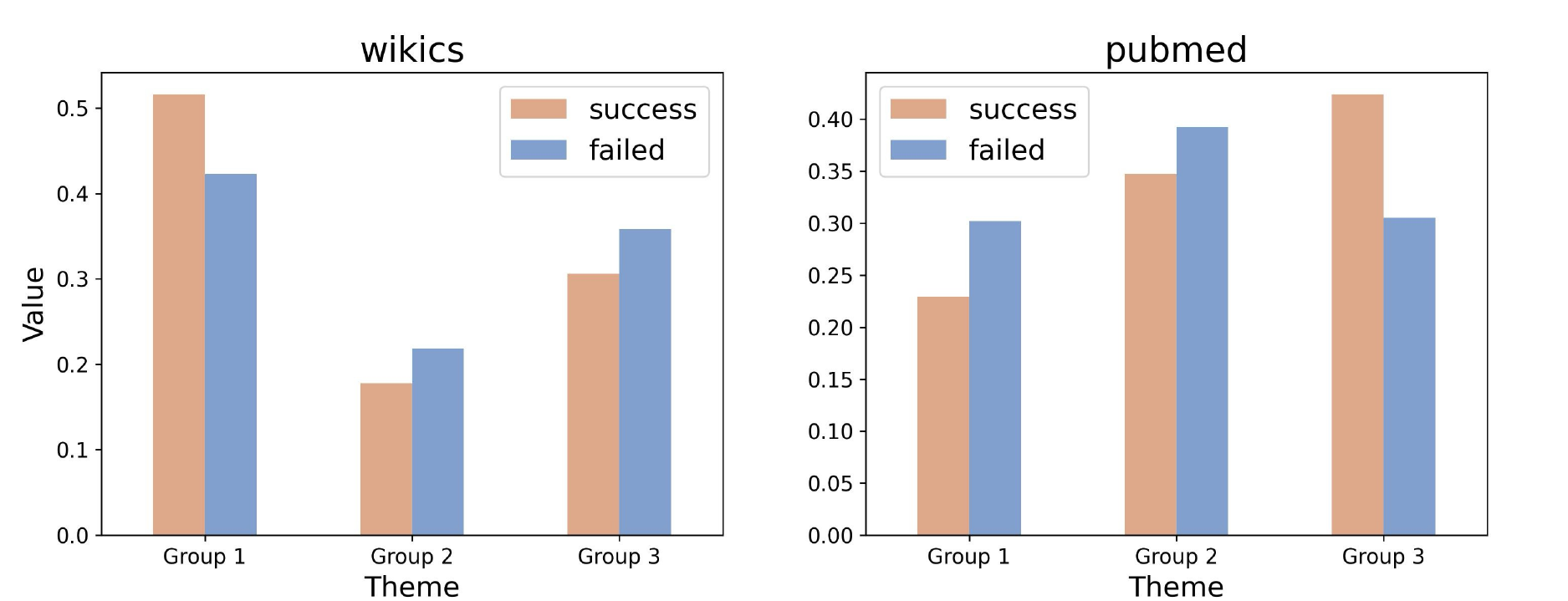} 
  \caption{Theme distributions of the node being attacked successfully and unsuccessfully.}
  \label{fig:analysis_LDA}
\end{figure}
\section{Combining Textual and Structural Attack}
\label{sec:sup-combine}
To enhance attack capabilities, maybe a combined framework that perturbs both text attributes and graph structure is needed. In this study, we provide some preliminary results. First of all, we design a simple strategy that combines the structural attack and textual attack directly, further improving the attack ability. The results are reported in Table~\ref{tab:combine}. In addition, we design various combination strategies for modifying the textual attack. These include prioritizing the attack on text attributes of small-degree nodes, targeting large-degree nodes first, and attacking text attributes within the same cluster. Based on the results in Table~\ref{tab:combine+}, we find that prioritizing attacks on the text attributes of small-degree nodes is more effective. This is merely a preliminary attempt, and we hope it will inspire more in-depth research.

% as shown in Fig.~\ref{fig:combined_attack_framework},
% \begin{figure}[h]
%   \centering
% \includegraphics[width=1\columnwidth]{figures/combined_attack.pdf} 
%   \caption{Combined attack framework.}
%   \label{fig:combined_attack_framework}
% \end{figure}

\begin{table}[tp]
\centering
\caption{Result of combined attack using SemAttack and PRBCD. Here we perform \textbf{5\%} perturbation on both text features and the graph structure.}
\label{tab:combine}
\resizebox{1\textwidth}{!}{%
\begin{tabular}{l|cc|cc|cc|cc}
\toprule
& \multicolumn{2}{c|}{Cora}      & \multicolumn{2}{c|}{Citeseer}  & \multicolumn{2}{c|}{PubMed} & \multicolumn{2}{c}{WikiCS}   \\ \midrule
Feature \textbackslash{} Ptb. 
& \multicolumn{1}{c|}{5\% edge} & \multicolumn{1}{c|}{5\% edge \& node} & \multicolumn{1}{c|}{5\%edge} & \multicolumn{1}{c|}{5\% edge \& node} & \multicolumn{1}{c|}{5\%edge} & \multicolumn{1}{c|}{5\% edge \& node} & \multicolumn{1}{c|}{5\%edge} & \multicolumn{1}{c}{5\% edge \& node} \\ \hline

BOW    
& \multicolumn{1}{c|}{78.40±1.67} & \multicolumn{1}{c|}{78.23±1.70}       & \multicolumn{1}{c|}{69.64±0.98} & \multicolumn{1}{c|}{69.58±1.18}       & \multicolumn{1}{c|}{74.18±0.76} & \multicolumn{1}{c|}{74.01±0.78}  & \multicolumn{1}{c|}{66.38±4.65} & \multicolumn{1}{c}{66.07±4.63}     \\

Sbert  
& \multicolumn{1}{c|}{79.93±0.49} & \multicolumn{1}{c|}{79.75±0.54}       & \multicolumn{1}{c|}{74.16±1.23} & \multicolumn{1}{c|}{74.15±1.31}       & \multicolumn{1}{c|}{75.92±1.40} & \multicolumn{1}{c|}{75.87±0.90}  & \multicolumn{1}{c|}{66.94±2.54} & \multicolumn{1}{c}{66.73±2.53}     \\

% E5     
% & \multicolumn{1}{c|}{81.38±1.52} & \multicolumn{1}{c|}{81.53±1.46}       & \multicolumn{1}{c|}{73.86±0.64} & \multicolumn{1}{c|}{73.70±0.73}       & \multicolumn{1}{c|}{79.27±1.45} & \multicolumn{1}{c|}{79.07±1.54}  & \multicolumn{1}{c|}{67.21±1.68} & \multicolumn{1}{c}{67.24±1.42}    \\

LLaMA  
& \multicolumn{1}{c|}{80.03±0.50} & \multicolumn{1}{c|}{79.14±1.17}  & \multicolumn{1}{c|}{71.70±1.50} & \multicolumn{1}{c|}{70.68±1.70}  & \multicolumn{1}{c|}{75.78±1.33} & \multicolumn{1}{c|}{74.95±0.98}  & \multicolumn{1}{c|}{66.60±5.54} & \multicolumn{1}{c}{66.41±5.64}     \\ 
\bottomrule
\end{tabular}}
\end{table}

\begin{table}[!th]
\centering
\caption{Results of \textbf{5\%} edge and \textbf{15\%} perturbations under three different sampling strategies on Cora. Small degree first and Large degree first means we sample nodes with the smallest or largest degrees respectively. Clustering refers to sampling nodes that are in the same cluster.}
\label{tab:combine+}
\resizebox{0.5\textwidth}{!}{%
\begin{tabular}{l|c|c}
\toprule
Cora Dataset                            & \multicolumn{2}{c}{5\% edge \& 15\% node} \\ \midrule
sampling\textbackslash{}perturbation    & ASR                 & ACC                 \\ \hline
Small degree first                    & \multicolumn{1}{c|}{17.20±3.33}       & \multicolumn{1}{c}{79.37±0.53}          \\
Large degree first                    & \multicolumn{1}{c|}{3.27±1.15}        & \multicolumn{1}{c}{79.44±0.85}          \\
Clustering                            & \multicolumn{1}{c|}{4.40±1.69}        & \multicolumn{1}{c}{79.53±0.61}  \\
\bottomrule
\end{tabular}}
\end{table}

\section{Broader Impact}
\label{sec:sup-impact}
Graph machine learning methods with Large Language Models (Graph-LLMs) have reported promising performance, covering a wide range of applications. We are initiating this benchmark to call more researchers' attention to whether Graph-LLMs are robust against graph adversarial attacks. The proposed benchmark can significantly promote the development of robustness in Graph-LLMs. Our benchmark provides a comprehensive evaluation, including LLM-as-Enhancers and LLM-as-Predictors, and consider both structural attacks and textual attacks. However, there are still many research gaps to be filled. Firstly, we have only explored node classification tasks and have yet to explore link prediction and graph classification tasks. Secondly, we have focused only on evasion attacks and have not addressed poisoning attacks. These limitations will be addressed in our future work.
% ------------------------------------------

\end{document}